\documentclass[10pt,twocolumn,letterpaper]{article}

\usepackage[pagenumbers]{cvpr} 

\usepackage{ifthen}
\usepackage[linesnumbered,ruled,vlined]{algorithm2e}
\usepackage{color}
\usepackage{xcolor}

\SetKwInput{KwInput}{Input}                
\SetKwInput{KwOutput}{Output}              

\usepackage{multirow}

\usepackage{makecell}
\usepackage{adjustbox}

\usepackage{bbm}
\usepackage{float}
\usepackage{alltt}
\usepackage{newlfont} 
\usepackage{array}
\usepackage{wrapfig}
\usepackage{booktabs}
\usepackage{amsfonts}
\usepackage{dsfont}
\usepackage{mleftright}

\usepackage{soul}
 
\usepackage{epsfig}
\usepackage{graphicx}
\usepackage{amsmath}
\usepackage{amssymb}

\definecolor{DeltaColor}{rgb}{0.039,0.73,0.71}
\definecolor{SetaColor}{rgb}{0.867, 0.0235, 0.376}
\definecolor{SigmaColor}{rgb}{0.98,0.45,0.0}
\definecolor{RedColor}{rgb}{0.8,0,0}
\definecolor{AlphaColor}{rgb}{0,0,0.8}
\definecolor{BetaColor}{rgb}{0.8,0,0.8}
\definecolor{GammaColor}{rgb}{0.5,0,0.7}
\definecolor{EpsilonColor}{rgb}{0.353,0.725,0.906}
\definecolor{TauColor}{rgb}{0.423,0.235,0.192}

\newcommand{\shortname}{AlignSAM}
\newcommand{\bsegname}{semantic recalibration module}

\definecolor{cvprblue}{rgb}{0.21,0.49,0.74}
\definecolor{kellygreen}{rgb}{0.3, 0.73, 0.09}
\definecolor{alizarin}{rgb}{0.82, 0.1, 0.26}

\newcommand{\xmark}{{\color{alizarin} \ding{55}}}
\usepackage[pagebackref,breaklinks,colorlinks,citecolor=cvprblue]{hyperref}
\usepackage{pifont}
\newcommand{\cmark}{{\color{kellygreen} \ding{51}}}
\usepackage{color}
\usepackage{xcolor}

\def\paperID{2058} 
\def\confName{CVPR}
\def\confYear{2024}

\title{AlignSAM: Aligning Segment Anything Model to Open Context via Reinforcement Learning} 

\author{Duojun Huang\textsuperscript{1,2}\quad Xinyu Xiong\textsuperscript{1}\quad Jie Ma\textsuperscript{1}\quad Jichang Li\textsuperscript{1,3}\quad Zequn Jie\textsuperscript{4}\quad Lin Ma\textsuperscript{4}\quad Guanbin Li\textsuperscript{1,2}\footnotemark[2]\\
\textsuperscript{1}School of Computer Science and Engineering, Sun Yat-sen University, Guangzhou, China \\
\textsuperscript{2}GuangDong Province Key Laboratory of Information Security Technology \\
\textsuperscript{3}The University of Hong Kong \quad \textsuperscript{4}Meituan
}

\begin{document}
\maketitle

\renewcommand{\thefootnote}{\fnsymbol{footnote}}
\footnotetext[2]{Corresponding author is Guanbin Li. This work was supported in part by the National Natural Science Foundation of China (NO.~62322608), and in part by the Open Project Program of the Key Laboratory of Artificial Intelligence for Perception and Understanding, Liaoning Province (AIPU, No.~20230003).}

\begin{abstract}
Powered by massive curated training data, Segment Anything Model (SAM) has demonstrated its impressive generalization capabilities in open-world scenarios with the guidance of prompts. 
However, the vanilla SAM is class-agnostic and heavily relies on user-provided prompts to segment objects of interest. 
Adapting this method to diverse tasks is crucial for accurate target identification and to avoid suboptimal segmentation results. In this paper, we propose a novel framework, termed \shortname{}, designed for automatic prompting for aligning SAM to an open context through reinforcement learning. Anchored by an agent, \shortname{} enables the generality of the SAM model across diverse downstream tasks while keeping its parameters frozen. 
Specifically, \shortname{} initiates a prompting agent to iteratively refine segmentation predictions by interacting with the foundational model. It integrates a reinforcement learning policy network to provide informative prompts to the foundational model. Additionally, a semantic recalibration module is introduced to provide fine-grained labels of prompts, enhancing the model's proficiency in handling tasks encompassing explicit and implicit semantics. Experiments conducted on various challenging segmentation tasks among existing foundation models demonstrate the superiority of the proposed \shortname{} over state-of-the-art approaches. Project page: \url{https://github.com/Duojun-Huang/AlignSAM-CVPR2024}. 

\end{abstract}

\section{Introduction}\label{sec:intro}

Compared to traditional computer vision tasks~\cite{li2019relation, wu2019enhancing, wu2019mutual, huang2023divide, jiaming2024learning}, 
segmentation stands as a fundamental task, playing a pivotal role in visual understanding systems. According to different semantic criteria for grouping pixels, various downstream segmentation tasks have emerged, such as saliency detection~\cite{CVPR20_MSNet,ICCV19_EGNet, CVPR23_MESOD}, shadow detection~\cite{CVPR19_DISTRACTION,TIP21_RevisitShadow}, and glass-like object detection~\cite{TIP22_ProgreGlassSeg,ICCV21_EnhanceGlassSeg}. 
Although significant progress has been achieved, the development of a unified framework that can accommodate the wide variations inherent in the formulations of diverse segmentation tasks continues to pose a challenge.

\begin{figure}[t]
    \centering    \includegraphics[width=1.0\linewidth]{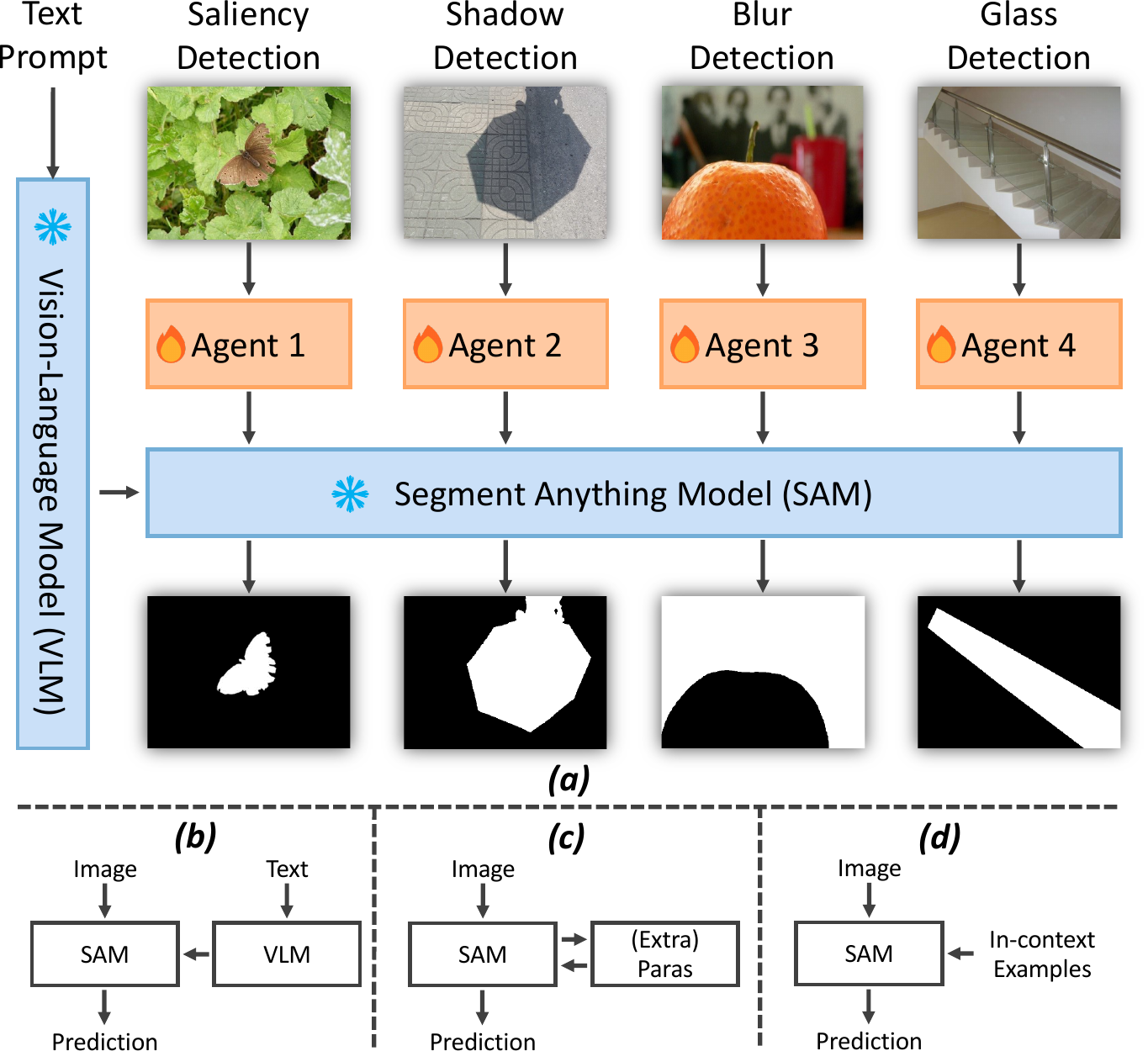}
    \caption{
    Conceptual comparisons of our method and previous approaches.  
    Frozen and learnable parameters are highlighted in \textbf{\textcolor[HTML]{73AED7}{blue}} and \textbf{\textcolor[HTML]{FCAF7C}{orange}}, respectively. 
    (a) The proposed method.
    (b) Text-guided methods~\cite{arXiv23_RefSAM,arXiv23_SAMCLIP}. (c)  PEFT methods~\cite{ICCVW23_SAM-Adapter,arXiv23_CustomSAM}. (d) In-context learning methods~\cite{arXiv23_PerSAM,arXiv23_Matcher}.
    Observed that
    the proposed agent-based auto-prompting effectively grasps vision and linguistic cues, unleashing the potential of the foundation segmentation model in various contexts, such as saliency detection, shadow detection, blur detection, and glass detection. 
    } 
    \label{fig:title}
\end{figure}

Recent breakthroughs in Vision Foundation Models (VFMs) have demonstrated impressive capabilities in zero-shot segmentation within open scenarios~\cite{CVPR23_Painter,ICCV23_SAM,ICCV23_SegGPT,arXiv23_SEEM}. 
Powered by the large-scale high-quality training dataset, Segment Anything Model (SAM)~\cite{ICCV23_SAM} has demonstrated its powerful capacity to be generalized into different tasks and data distributions unseen during training. 
However, SAM heavily relies on manual prompts such as bounding boxes and points to segment objects of interest since it is pretrained in a class-agnostic manner.  
Consequently, achieving target-aware adaptation of SAM for different tasks plays a crucial role in bridging the gap between class-agnostic predictions and industrial needs. 

\begin{figure}[t]
    \centering
\includegraphics[width=1.0\linewidth]{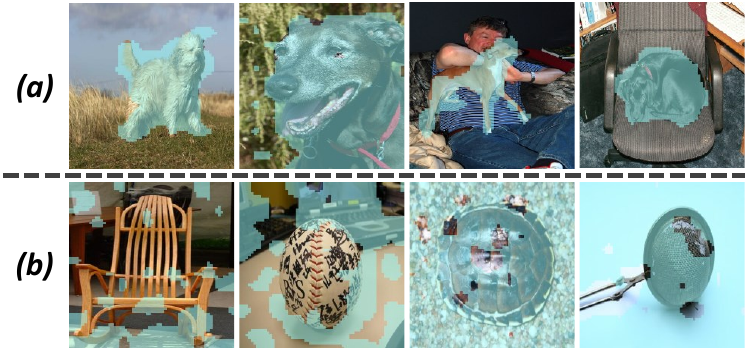}
    \caption{Segmentation results for CLIP-Surgery~\cite{arXiv2023_sclip} using different prompts. (a) and (b) illustrate the prompts of ``dog'' and ``salient object'', respectively. Observed that vision-language models excel in processing explicit semantics but often struggle with implicit semantics.} 
    \label{fig:semantic_level}
\end{figure}

Language-guided segmentation has emerged as a promising approach to customize SAM for specific downstream tasks. 
By leveraging the cross-modal alignment of the vision-language model~\cite{ICML21_CLIP}, referring segmentation can be accomplished by accurately depicting the foreground targets. Nevertheless, there are numerous visual targets that are difficult to accurately express by textual description. 
For example, the salient object detection task requires visual comparison among different regions within an image, which poses challenges in aligning it with the text encoder as evident in~Figure~\ref{fig:semantic_level}.

One feasible way to capture contextual information is to utilize sufficient training samples to fine-tune the foundation models. Owing to the substantial number of parameters in foundation models, conducting full fine-tuning for adaptation leads to a significant computational burden. 
Recent advancements have been directed toward developing Parameter-Efficient Fine-Tuning (PEFT) methods. These methods integrate adapters~\cite{ICML19_Adapter,ICCVW23_SAM-Adapter} or LoRA~\cite{ICLR22_LoRA} blocks while keeping the backbone network frozen, as indicated in Figure~\ref{fig:title}(c). They aspire to attain promising performance comparable to the paradiam of full finetuning. Nevertheless, PEFT methods necessitate the calculation of gradients in the intermediate layers of the backbone network, which may be inaccessible due to privacy concerns. Furthermore, these methods remain highly reliant on an abundance of training data. 
To achieve sample-efficient adaptation, recent researches~\cite{arXiv23_PerSAM,arXiv23_Matcher} explore in-context learning by leveraging similarity matching to generate point prompts for personalized adaptation using limited training instances. However, similarity calculations assume  are prone to interference, impeding the capture of implicit semantics. 

Inspired by the model-free spirit of reinforcement learning (RL), we introduce a unified framework for adapting SAM to diversified scenarios while keeping parameters of the backbone network frozen. 
By formulating automatic prompting as a sequential decision-making process, an agent is trained to imitate human annotators to recommend prompting positions. The RL framework is built by an actor branch and a critic branch networks, which collaborate to learn the optimal prompting policy to efficiently refine the segmentation outputs. 
Moreover, a semantic recalibration module is additionally introduced to provide reliable labels for the selected positions. It contributes to delineating foreground and background areas for downstream tasks of different types. Concretely, for tasks with explicit semantic, a cross-modal attention module integrates visual and linguistic information from the context. In tasks with implicit semantics, a visual-similarity learning branch is presented to capture implicit contextual concepts. 

The contributions can be summarized as follows, with Figure~\ref{fig:title}(a) showcasing its conceptual illustration. 
\begin{itemize}
    \item 
    A general approach, termed AlignSAM, is proposed to optimize the automatic prompting policy for efficiently adapting foundation model to downstream tasks. By constructing universal actions, state and reward signals, it is able to handle various types of downstream tasks within a unified framework.

    \item 
    A semantic recalibration module is introduced to provide precise prompting labels for adapting the vision foundation model to tasks with explicit and implicit semantics. 
    
    \item Experiments conducted on various challenging segmentation tasks among existing foundation models demonstrate the superiority of the proposed method over state-of-the-art approaches for efficient adaptation. 
\end{itemize}

\section{Related Work}

\paragraph{Image segmentation.} 
Different from traditional tasks~\cite{li2019semi, li2021cross, li2022nce, li2023idm, li2023betweenness, li2024feddiv, zhang2024distribution}, 
leveraging richly annotated datasets~\cite{IJCV10_PASCALVOC,CVPR16_CITYSCAPE} and advanced pretrained feature extractors~\cite{CVPR16_ResNet,ICLR15_VGG,zhang2024masked}, deep image segmentation has yielded notable results in diverse applications~\cite{TPAMI21_SegSurvey1,xiong2023unpaired,li2023hybridvps,ma2023enhanced,10097456}, primarily through the development of complex network architectures and training methodologies. For example, DeepLabV3Plus~\cite{ECCV18_DeepLabV3} in semantic segmentation utilizes a spatial pyramid pooling structure for multiscale contextual information. In shadow detection, FDRNet~\cite{iccv21_fdrnet} enhances training samples by adjusting brightness to differentiate between dark non-shadow and bright shadow areas. Meanwhile, for saliency detection, MENet~\cite{CVPR23_MESOD} proposes a method for progressively aggregating and refining features to address accuracy challenges in cluttered scenes. However, these specialized methods lack versatility, limiting their transferability to different segmentation tasks.

\paragraph{Vision foundation models.} 
Leveraging large-scale datasets such as~\cite{laion400m,nips22_laion5b}, vision foundation models have made significant advancements in various tasks of computer vision, encompassing image classification~\cite{ICML21_CLIP, icml22_align}, segmentation~\cite{ICCV23_SAM,arXiv23_SEEM,ICCV23_SegGPT}, and generation~\cite{cvpr22_ldm}. CLIP~\cite{ICML21_CLIP}, trained on massive image-caption pairs, exhibits robust zero-shot learning capabilities in cross-modal task. Segment Anything Model~\cite{ICCV23_SAM} introduces a large-scale segmentation dataset SA-1B and a training framework to enable prompt-driven segmentation in a zero-shot manner.
Painter~\cite{CVPR23_Painter,ICCV23_SegGPT} integrates multiple segmentation tasks within an in-context learning framework. 
Despite their effectiveness, it remains challenging for these models to adapt to diversified downstream tasks while preserving great interpretability as demonstrated in Sec.~\ref{sec:intro}.

\paragraph{Parameter-Efficient Fine-Tuning.}
Foundation models inherently possess a massive amount of parameters. Therefore, directly fine-tuning all of these parameters for each downstream task results in significant computational costs, which limits the efficiency of specialized applications. 
Recent efforts have been made to optimize extra learnable parameters on a small-scale, rather than those of the backbone network. Existing approaches for efficient fine-tuning of visual parameters generally fall into three categories: prompt tuning~\cite{ECCV22_VPT}, adapter methods~\cite{ICML19_Adapter,MLMI23_MammoSAM,ICCVW23_SAM-Adapter}, and low-rank adaptation (LoRA)~\cite{ICLR22_LoRA}.  
Specifically, VPT~\cite{ECCV22_VPT} injects a small number of learnable parameters into the Transformer's input space and keeps the backbone frozen during the fine-tuning stage. 
Adapter-based techniques~\cite{ICML19_Adapter,MLMI23_MammoSAM,ICCVW23_SAM-Adapter}, such as SAM-Adapter~\cite{ICCVW23_SAM-Adapter}, integrate tunable modules into pre-trained large models to suit domain-specific tasks, where MLP layers are inserted into each transformer block of the image encoder. 
Furthermore, LoRA~\cite{ICLR22_LoRA} hypothesizes that the change matrix of weights during adaptation has a low rank, and therefore only tunes a pair rank decomposition matrix of the original linear layers weights, while keeping the pre-trained weights frozen.
However, most of these algorithms heavily rely on sufficient training data during adaptation and require gradient calculation of the intermediate layers of the foundation model, resulting the lack of sample-efficiency and model-agnostic property. 

\paragraph{Reinforcement learning for computer vision.} Reinforcement learning has been successfully applied to various vision scenarios including image classfication~\cite{nips14_RAM,ICCAS20_GRAM}, object tracking~\cite{iclm18_rltrack1, arxiv17_rltrack2} and semantic segmentation~\cite{arxiv16_rlseg,iclr20_rlal}. Notably, RAM~\cite{nips14_RAM} formulated attention-based image processing as a sequential control problem and integrated reinforcement learning and recurrent neural network for object recognition task. The reinforced active segmentation method~\cite{iclr20_rlal} achieves comparable results with significantly less annotated data than other weakly supervised approaches. 
Inspired by the model-free sprit of reinforcement learning, we train a universal agent to efficently prompt the vision foundation model, constructing a unified framework that can scale to diversified downstream tasks. 

\section{Methodology}
In this section, we first present a revisit of the Segment Anything Model (SAM)~\cite{ICCV23_SAM} and then introduce the task to be addressed in this paper. 
Next, we delve into how to achieve the pipeline of the proposed framework and introduce the \bsegname{} algorithm. 
Finally, we outline the training and evaluation processes employed in our approach. 

\begin{figure*}[t]
    \centering
    \includegraphics[width=.8\linewidth]{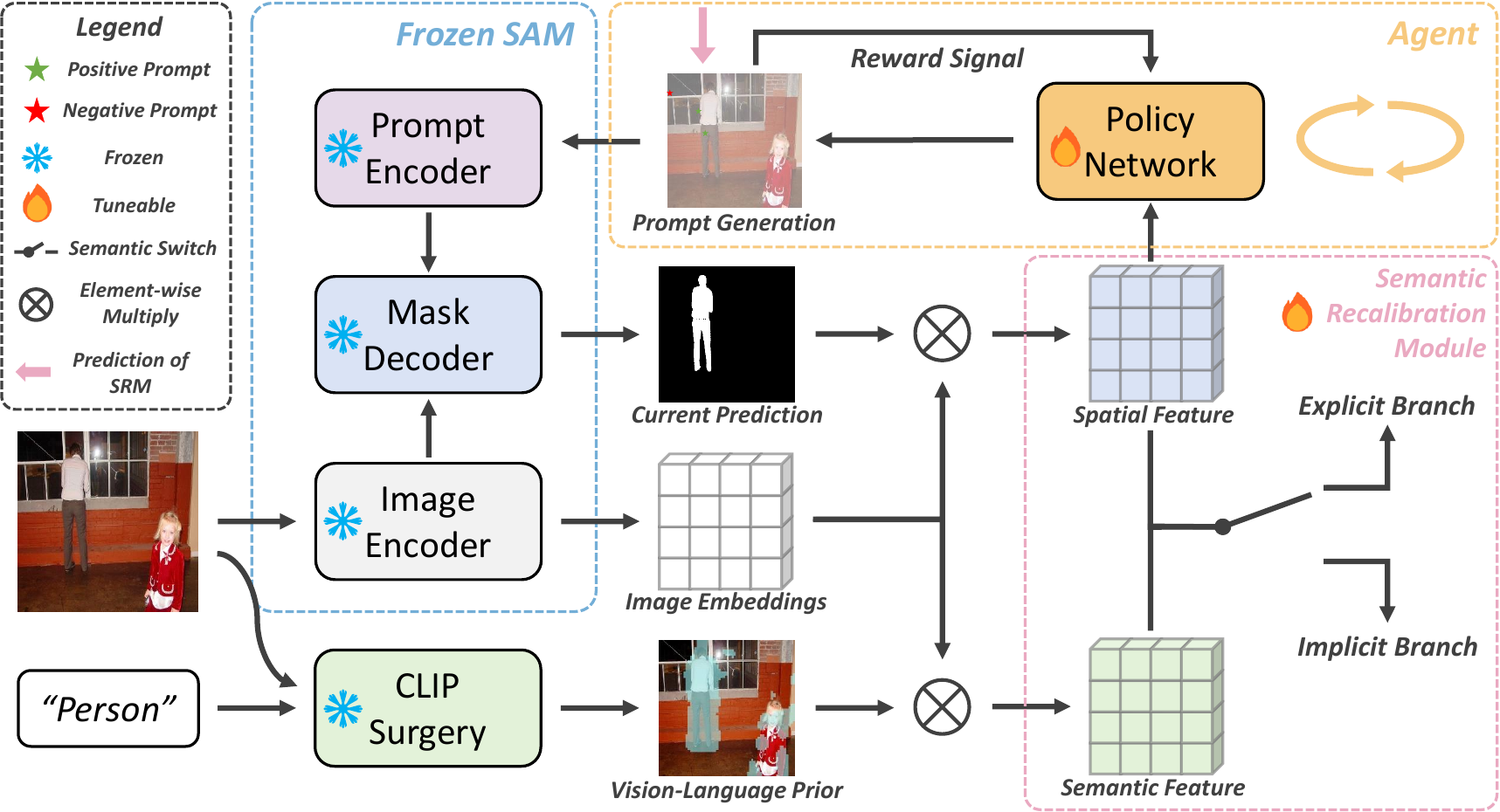}
    \caption{An overview of the proposed AlignSAM, which consists of a Segment Anything Model (SAM), a vision-language model (CLIP-Surgery~\cite{arXiv2023_sclip}), a reinforcement learning agent, and a semantic recalibration module. 
    The frozen SAM receives the point prompts generated by the agent and semantic recalibration module, dealing with various downstream scenarios without relying on manual prompting. 
    }    
    \label{fig:method}
\end{figure*}

\subsection{Preliminary}
\label{sec:prob_form}

\paragraph{Revisit of Segment Anything Model.} 

The Segment Anything Model (SAM) is proposed as a vision foundation model for achieving a unified capacity of segmentation. In this case, given a set of prompts, i.e., foreground and background points, bounding boxes, or a mask, SAM is capable of segmenting the prompted target. In detail, the SAM model is made up of an image encoder $\mathcal{E_I}(\cdot)$, a prompt encoder $\mathcal{E_P}(\cdot)$ and a mask decoder $\mathcal{D_M}(\cdot)$. During model inference, the image encoder and the prompt encoder are first utilized to encode the given image $x$ and a prompt set $P$ respectively, which can be formulated as follows: 
\begin{equation}
   F_I = \mathcal{E_I}(x), ~~~P_t = \mathcal{E_P}(P), 
\end{equation}
where $F_I\in R^{h \times w \times c}$ and $P_t\in R^{k \times c}$, with $k$ denoting the amount of prompts. Further, the image feature $F_I$ and prompt embedding $P_t$ will be fed into $\mathcal{D_M}(\cdot)$ to generate the prediction mask $M$ as follows: 
\begin{equation}
   M = \mathcal{D_M}(F_I, P_t).
\end{equation}

\noindent
\textbf{Overview.}
In this paper, we present a novel framework termed \shortname{} for adapting SAM to diversified scenarios while keeping its parameters frozen. First, we formulate automatic prompting as a sequential decision-making process and introduce Target-aware Reinforcement Learning to construct a unified pipeline to handle tasks with diversified objects. Specifically, we train a reinforcement agent to imitate human annotators to recommend prompting positions, which executes a series of prompting actions to refine the segmentation progressively. Moreover, a Semantic Recalibration Module is additionally proposed to delineate foreground and background areas,  providing precise label for the selected prompts. 

\subsection{Target-aware Reinforcement Learning} 
\label{sec:rl}
We train a target-aware reinforcement agent to interact with the environment to maximize the accumulated rewards. 
The objective of the agent is to recommend optimal prompting positions to facilitate progressive refinement of the mask prediction by the segmentation model. To be specific, during each interaction loop, the agent will first select an action according to current state observation and subsequently receive a signal of reward from the environment. 
Here, we first present how to construct the state space, action space, and the reward function separately in our RL framework, and then illustrate the process of model training. 
\vspace{-5mm}

\paragraph{Action space.} To construct the action space, a naive approach is to regard each pixel as a candidate action in the image. However, it is cost intensive and in-efficient. 
To relieve the computation burden while maintaining the local details of image, we propose to construct action space in the patch-level rather than pixel-level for each image. 
Specifically, we propose to divide each image into different regions to build the action space for reinforced learning. Given an input image $x \in R^{H\times W \times C}$, we divide it into image patches $x_p \in R^{H' \times W' \times C}$. The candidate action set is the center points of each patch in an image, which can be formulated as follows:
\begin{equation}
A = \{(h_i, w_i)|(h_i, w_i)=center(x_p)\}.   
\end{equation}
Therefore, the action set consists of $N=\frac{H}{H'}\cdot \frac{W}{W'}$ candidate points in an image. For each image, the goal is to select the optimal action among the $N$ candidates at each time-step, and the process will be terminated after $T$ time-steps in total. 

\vspace{-5mm}
\paragraph{State space.} 
At each timestep, the agent determines which action to execute based on the current state. In order to facilitate the agent to perform gainful decisions, the state must encompass comprehensive information about the environment. Therefore, we combine the visual feature embedding and the iterative prediction to form the state representation. For an image sample $x$ at the timestep $t$, the state representation can be formulated as: 
\begin{equation}
\label{state}
   s_t = \mathcal{E_I}(x) \cdot M_{t-1},  
\end{equation}

where $M_{t-1}$ denotes the mask predicted by the mask decoder of SAM at last step. 
The feature extracted by the image encoder represents the inherent spatial feature of the sample, while the information derived from the previous prediction maintains the contextual continuity across the sequential dimension. The multiplication operation highlights the features of regions that are more likely to be recognized as foreground in the historical iteration round and diminishes the features of regions that are prone to be classified as background. In this way, the policy network maps the state to a probability distribution over the actions, enabling the optimization of the maximum expected reward in the full training process. 

\paragraph{Reward function.}
Previous RL-based methods for computer vision tasks, such as~\cite{iclr20_rlal, a3rl}, defined the reward function to improve the task performance between iterations. Working as a binary segmentation task, the performance metrics are diversified across segmentation scenarios. To relieve the burden of hand-crafted function engineering, we formulate the reward function as the score of querying the target foreground. 
At each timestep, if the agent's action points to the foreground region, it earns a reward of \textbf{+1}; otherwise, it incurs a penalty of \textbf{-1} as feedback.

\paragraph{Training RL model.}
We employ a widely-used proximal proxy optimization (PPO) algorithm~\cite{arxiv17_ppo} to instantiate the RL framework, which consists of an actor network $\pi_\theta$ and a critic network $\mathcal{V}_\theta$. The actor explores and improves its policy by interacting with the environment by mapping the state to a probability distribution over all the actions. Meanwhile, the critic is trained in a supervised manner to predict the state value of the current policy. The actor and critic work in a cooperative manner to learn the optimal policy.

During the training phase, the RL networks will be trained by $E$ episodes with each episode containing $T$ prompting steps. At the beginning of each episode, the prompt set $P_0$ for each image is initialized as a pair of positive and negative prompt points randomly selected from the sample itself. After the first inference prompted by $P_0$, the initial state $s_0$ can be calculated according to Eq.~\ref{state}. At each prompting step $t$, the agent selects an action from the probability distribution predicted by the actor network, which can be formulated as follows: 
\begin{equation}
    a_t\sim p_t(\cdot)=\pi_{\theta}(\cdot|s_t), 
\end{equation}
where $p_t$ represents the probability distribution across the action space. After action $a_t$ is executed, the agent receives a scalar reward $r_{t}$ from the environment. At each iteration, the tuple comprising the current state, action, reward, and next state is stored in a memory buffer. When an episode is completed, the parameters of $\pi_\theta$ and $\mathcal{V}_\theta$ will be updated using stochastic gradient descent (SGD) for $K$ epochs. To form the optimization target, the action value function is first calculated as follows:
\begin{equation}
\begin{split}
Q(s_t, a_t) = r_t + \gamma r_{t+1} + ... + \gamma^{T-t} \mathcal{V}_\theta(s_T), 
\end{split}
\end{equation}
where $\gamma$ denotes a discount factor to balance between the current reward and future expected values. The action value represents the expected cumulative reward after taking action $a_t$ under state $s_t$. Furthermore, the advantage function can formulated as follows: 
\begin{equation}
\begin{split}
A_t = Q(s_t, a_t) - \mathcal{V}_{\theta}(s_t), 
\end{split}
\end{equation}
where $\mathcal{V}_{\theta}(s_t)$ denotes the state value estimated by the critic network. The advantage function measures the expected value of taking specific action under a given state over the average performance. To improve the training stability, importance resampling is utilized during the optimization of policy model. Specifically, the probability ratio $\gamma_\theta(t)$ between the previous and current actor models is formulated as follows: 
\begin{equation}
\gamma_\theta(t)=\frac{\pi_\theta\left(a_t \mid s_t\right)}{\pi_\theta\left(a_{t-1} \mid s_{t-1}\right)}.
\end{equation}

\begin{algorithm}[t]
\DontPrintSemicolon
    \KwInput{Input image $x$ and its ground truth mask $y$, SAM's image encoder $\mathcal{E_I}(\cdot)$, prompt encoder $\mathcal{E_P}(\cdot)$, mask decoder $\mathcal{D_M}(\cdot)$ } 
   \KwOutput{Optimal parameters of actor network $\pi_{\theta}$ and critic network $\mathcal{V}_{\theta}$ } 
    
    \For{$e=1,2,...,E$}{
        $P_0= \{a_p, a_n\}\sim y$ 
        
        $M_0= \mathcal{D_M}(\mathcal{E_I}(x), P_0) $
        
        $s_1 = \mathcal{E_I}(x) \cdot M_0. $
        
        \For{$t=1,2,...,T$}{
        $a_t\sim \pi_{\theta}(\cdot|s_t)$
        
        $P_t = P_{t-1}\cup \{a_t\}$
        
        $M_t= \mathcal{D_M}(\mathcal{E_I}(x), P_t) $
        
        $s_t = \mathcal{E_I}(x) \cdot M_t. $
        
        
        $r_{t}=$ \textbf{+1} if $y(a_t)$=1 else \textbf{-1}   
        
        }
        // Compute advantage estimates
        
        \For{$t=T,T-1,...,1$}{
        $\hat{A}_t = Q(s_t, a_t) - \mathcal{V}_{\theta}(s_t) $ 
        }
        
        // Update $\pi_{\theta}$ and $\mathcal{V}_{\theta}$ with $K$ epochs 
        
        \For{$k=1,2,...,K$}{
        
        $\pi_{\theta} \leftarrow \pi_{\theta} + \nabla_{\pi_{\theta}}L_{act} $ 
        
        $\mathcal{V}_{\theta} \leftarrow \mathcal{V}_{\theta} - \nabla_{\mathcal{V}_{\theta}}L_{cri} $    
        }
    }

\caption{Training procedure of the RL model. }
\label{algo:RL Training}
\end{algorithm}

With the advantage function denoted by $A_{t}$ at timestep $t$, the objective function of the actor network is formulated as follows: 
\begin{multline}
      L_{act} = \mathop{\mathds{E}}\limits_{ t\in [1, T] } \big [ min(\gamma_{\theta}(t)A_{t}, clip(\gamma_{\theta}(t),1-\epsilon, 1 + \epsilon)A_{t} \big ], 
\end{multline}
where $clip(\gamma_{\theta}(t),1-\epsilon, 1 + \epsilon)$ is a clipping function to constrain the value of $\gamma_{\theta}(t)$ to the interval between $(1 - \epsilon, 1 + \epsilon)$, and $\epsilon$ is a hyperparameter to control the magnitude of model update. 
Concurrently, the critic network is optimized to minimize the discrepancy between the estimated and the actual values. The training objective of the critic network can be formulated as follows: 
\begin{equation}
L_{cri} =  \mathop{\mathds{E}}\limits_{t\in [1, T] } \big [ ( Q(s_t, a_t) - \mathcal{V}_{\theta}(s_t))^2 \big ].
\end{equation}
The training process of the RL model is summarized in Algorithm~\ref{algo:RL Training}.

\subsection{Semantic Recalibration Module}
After tuning with reward, the reinforcement learning agent tends to prompt positions of foreground target for the segmentation model. A naive prompting policy is to regard all the selected points as target foreground. 
However, it is unrealistic and overly idealistic due to the limited capacity of the action space. 
To generate prompts with reliable label information, we design a semantic recalibration module (SRM) to transition the prompting from a coarse-grained policy to a fine-grained one. 
This module can generalize to tasks with explicit and implicit semantics by switchable branches. Eventually, the SRM output serves as the reference mask for the selected action to construct the complete prompt information. 

Specifically, to enhance the RL agent's comprehension of contextual information, we propose to recalibrate the state representation by considering both the semantic-dominated and spatial-dominated aspects of the environment. 
Following prior research~\cite{arXiv2023_sclip}, we adopt an approach to encode explicit semantic context using the text-guided attention map derived from the visual-language model CLIP~\cite{ICML21_CLIP}, achieved without requiring additional training. The attention map, denoted as $M_c$, is derived from the pairwise similarity between the image feature and the corresponding text feature extracted by the visual-language model. As for the spatial-dominated state, we employ the previous probability map predicted by the mask decoder of SAM to recalibrate the state, as it contains historical information learned from the visual samples. Finally, the semantic-dominated and the spatial-dominated states are formulated as the multiplication between their corresponding attention masks and the inherent feature of the sample, which can be formulated as follows: 
\begin{equation}
    s_c = \mathcal{E_I}(x) \cdot M_c, \\
    s_t = \mathcal{E_I}(I) \cdot M_{t-1}.
\end{equation}
Subsequently, the SRM receives states representation as input to predict the prompting mask for each candidate position, which can be formulated as: 
\begin{equation}
\label{GAP}
   y_r = K(V(s_c, s_t)), 
\end{equation}
where $K(\cdot)$ represents the classification head and $V(\cdot, \cdot)$ is a semantic switch which chooses the implicit branch or explicit branch to execute, depending on the type of downstream task. 

\paragraph{Implicit branch.}
The implicit branch is designed for segmentation tasks with abstract concept such as saliency detection, where the text feature is unavailable, since the prompt ``salient object'' is implicit semantics, as observed in Figure~\ref{fig:semantic_level}. Concretely, the $ V(\emptyset, s_t)$ is utilized in Eq.~\ref{GAP} and formulated as two convolution blocks to extract the targeted spatial feature without noisy attention from the text feature. 

\paragraph{Explicit branch.}
The explicit branch is designed for concrete object segmentation. Since textual information allows for a more comprehensive understanding of the concrete content in the image, we utilize $V(s_c, s_t)$ and formulate it as three convolution layers to aggregate the two states followed by a self-attention layer to capture the relationships and dependencies between different elements in an image. Finally, the feature is sent to two convolution layers to conduct the mask decoding. 

\paragraph{Training objective.}
The training objective to optimize the proposed implicit and explicit branches is a combination of the Dice loss~\cite{ICDM20_Dice} and the binary cross-entropy loss~\cite{ECCV18_DeepLabV3}, which are both widely adopted in image segmentation: 
\begin{equation}
L_{seg} = L_{dice}(y_r, y') + L_{bce}(y_r, y'), 
\end{equation}
where $y'$ denotes the ground truth mask of such a sample $x$ downsampled to the same size of $y_r$. During training, the SRM module is updated over $q$ iterations after the execution of the action. During evaluation, the prompt set should be initialized as a positive point and a negative point respectively responsible for the predicted probabilities with the highest and the lowest scores by the SRM module. 
During evaluation, the prompt set will be initialized as a positive point and a negative point with the highest and the lowest scores predicted as foreground areas by the SRM module. 

\section{Experiments}
\subsection{Experimental Setups} 
\paragraph{Datasets.} 

\begin{table}
\centering
\footnotesize
\begin{tabular}{lcc} 
\hline
Name & Task & Branch  \\ 
\hline
\textbf{CUHK}~\cite{ICCV14_CUHK} & Blur Detection  & I\\
\textbf{SBU}~\cite{ECCV16_SBU} & Shadow Detection  & E\\
\textbf{MSD}~\cite{ICCV19_MSD} & Glass Detection  & I\\
\textbf{DUTS}~\cite{CVPR2017_DUTS} & Saliency Detection & I\\
\textbf{PASCAL}~\cite{IJCV10_PASCALVOC} & Semantic Segmentation & E\\
\hline
\end{tabular}
\caption{Summary of datasets used in our benchmark. 
The term ``Branch'' refers to the semantic branch allocated for this dataset, with ``I'' representing the implicit branch and ``E'' denoting the explicit branch.
}
\label{tab:dataset}
\vspace{-5pt}
\end{table}
\begin{table*}[t]
\centering
\footnotesize
{
\begin{tabular}{l|cc|cc|cc|cc}
\toprule
\multirow{3}{*}{Method} & \multicolumn{2}{c|}{\textbf{Blur}} & 
\multicolumn{2}{c|}{\textbf{Shadow}}  & \multicolumn{2}{c|}{\textbf{Glass}} & \multicolumn{2}{c}{\textbf{Saliency}} \\
      &  \multicolumn{2}{c|}{CUHK~\cite{ICCV14_CUHK}} & \multicolumn{2}{c|}{SBU~\cite{ECCV16_SBU}}  & \multicolumn{2}{c|}{MSD~\cite{ICCV19_MSD}} & \multicolumn{2}{c}{DUTS~\cite{CVPR2017_DUTS}}\\ 
     &  mIoU $\uparrow$ &  $F_{\beta}\uparrow$   & mIoU $\uparrow$ & BER $\downarrow$  & mIoU $\uparrow$ &  $F_{\beta}\uparrow$ &  $E_\phi \uparrow$ & MAE $\downarrow$   
          \\ \hline 

SAMed~\cite{arXiv23_CustomSAM} & 55.44 & 71.68 & 17.24 & 42.84 & 41.35 & 52.67 & 76.41 & 0.104
 \\  
SEEM~\cite{arXiv23_SEEM}       & 60.84 & 67.45 & 19.52 & 48.01 & 32.35 & 37.44 & 59.88 & 0.326 
 \\ 
Painter~\cite{CVPR23_Painter} & 18.61 & 27.25 & 9.31 & 47.89 &  8.62 & 14.85 & \textbf{81.06} & 0.113 
 \\
PerSAM~\cite{arXiv23_PerSAM} & 55.84 & 71.59 & 18.68 & 49.51 & 31.18 & 38.00 & 64.13 & 0.257
\\ 

\textbf{Ours} & \textbf{68.47} & \textbf{76.89} & \textbf{30.78} & \textbf{34.62} & \textbf{45.44} & \textbf{57.28} & 78.21 & \textbf{0.082} \\ 
\hline
Ours-w/o RL & 59.75 & 70.98 & 25.62 & 37.57 & 33.41 & 46.91 & 74.19 & 0.086 \\ 
Ours-w/o SRM & 66.89 & 73.15 & 21.29 & 42.72 & 31.92 & 36.76 & 30.52 & 0.587 \\  
Ours-w/o MSI & 46.92 & 67.02 & 18.65 & 42.91 & 35.86 & 47.95 & 68.59 & 0.134 \\  
\bottomrule
\end{tabular}}
\caption{
Comparison results with existing SOTA approaches and ablation study results on four different segmentation tasks. The best performance among all approaches is highlighted in \textbf{blod}. ``RL", ``SRM", ``MSI" represent reinforcement learning, semantic recalibration module, multi-step interaction, respectively. 
}
\label{tab:sota_finetune}
\vspace{-2mm}
\end{table*}

\begin{table*}[t!]
\footnotesize
\centering
	\begin{tabular}{l|c|ccccccccccc}
	\toprule
  \makecell*[l]{Method} &\makecell*[c]{mIoU}
        &\makecell*[c]{Bottle}  &\makecell*[c]{Car}  &\makecell*[c]{Sheep} &\makecell*[c]{Cat}
        &\makecell*[c]{Chair}
        &\makecell*[c]{Dog}&\makecell*[c]{Person}&\makecell*[c]{Sofa}&\makecell*[c]{Cow} &\makecell*[c]{Horse}\\
		\cmidrule(lr){1-1} \cmidrule(lr){2-2}  
  \cmidrule(lr){3-12}
        MSA~\cite{arXiv23_MSA} & 47.12 & 31.56 & 41.05 & 56.47 & 60.65 & 19.79 & 55.95 & 38.81 & 33.97 & 60.38 & 61.99 \\   
        
        SAMed~\cite{arXiv23_CustomSAM} & 51.42 & 36.13 & 48.72 & 60.59 & 58.46 & 16.22 & 71.48 & 49.08 & 47.65 & 74.33 & 71.72 \\
        
        SEEM~\cite{arXiv23_SEEM}  & 52.30 & 36.95 & 44.95 & 63.93 & 84.46 & 19.67 & 66.40 & 63.87 & 47.89 & 73.74 & 74.10 \\
        
        Painter~\cite{CVPR23_Painter} & 59.27 & 35.33 & 61.93 & \textbf{77.29} & 63.16 & 29.90 & 59.82 & 46.40 & 55.58 & \textbf{78.71} & 72.82 \\
        
        PerSAM~\cite{arXiv23_PerSAM} & 53.02 & 36.03 & 46.94 & 64.42 & 69.39 & 22.28 & 67.25 & 49.07 & 36.84 & 68.79 & 65.56 \\

        \bf Ours & \textbf{62.09} & \textbf{48.09} & \textbf{66.13} & 73.12 & \textbf{85.99} & \textbf{31.68} & \textbf{79.84} & \textbf{64.72} & \textbf{61.63} & 72.97 & \textbf{75.10} \\
        \cmidrule(lr){1-12}
        Ours-w/o RL & 54.06 & 41.69 & 56.28 & 69.30 & 80.80 & 26.84 & 69.16 & 50.83 & 48.20 & 62.74 & 72.91 \\
        
        Ours-w/o SRM & 27.73 & 19.78 & 27.42 & 30.33 & 39.29 & 14.08 & 29.16 & 24.91 & 30.39 & 31.32 & 29.02 \\

	\bottomrule
	\end{tabular}
\caption{
Comparison results with existing SOTA approaches and ablation study results on Pascal-VOC 2012. 
We report the mean IoU of all 20 categories and the IoU of 10 randomly selected categories.
The best performance among all approaches is highlighted in \textbf{bold}. 
}
 \label{tab:pascal}
 \vspace{-4mm}
\end{table*}

We validate AlignSAM in various challenging benchmark datasets. The detailed dataset partition is shown in Table~\ref{tab:dataset}. 
For quantitative comparison, we follow the previously used evaluation metrics in each task, including mean intersection over union (mIoU), 
mean absolute error (MAE), balance error rate (BER)~\cite{ICCV15_BER}, F-measure ($F_\beta$)~\cite{CVPR09_Fmeasure}, and E-measure ($E_\phi$)~\cite{IJCAI18_Emeasure}. For each segmentation task, we randomly choose select 50 samples as the training set and utilize the original testing set for evaluation. 
 
\paragraph{Implementation details.} All the experiments are performed on a single NVIDIA A100 GPU with 80 GB memory. We export Adam as the optimizer conducted on all experiments, with a learning rate of $1.0\times10^{-4}$. To construct the action space, we set both height and width of each image to 800, and each patch within an image is assigned dimensions of 80 for both height and width. In the training of the RL agent, the hyperparameters $\gamma$ and $\epsilon$ are assigned the values of 0.99 and 0.20, respectively.
Additionally, we set the episode $E$ to 50 and set the epoch $K$ to 20. Furthermore, we fix the number of interaction rounds $T$ to 15. For the implicit branch, $Q$ is set to 1,  wherea for the explicit branch, it is set to 5.

\paragraph{Baselines.} We compare our proposed AlignSAM with following state-of-the-art (SOTA) approaches based on foundation models. Specifically, MSA~\cite{arXiv23_MSA} and SAMed~\cite{arXiv23_CustomSAM} are parameter-efficient fine-tuning methods based on adapter~\cite{ICML19_Adapter} and LoRA~\cite{ICLR22_LoRA}; Painter~\cite{CVPR23_Painter} and SEEM~\cite{arXiv23_SEEM} are other foundation segmentation models; PerSAM~\cite{arXiv23_PerSAM} is an in-context learning variant of SAM.

\begin{figure}[t]
    \centering
    \includegraphics[width=1.0\linewidth]{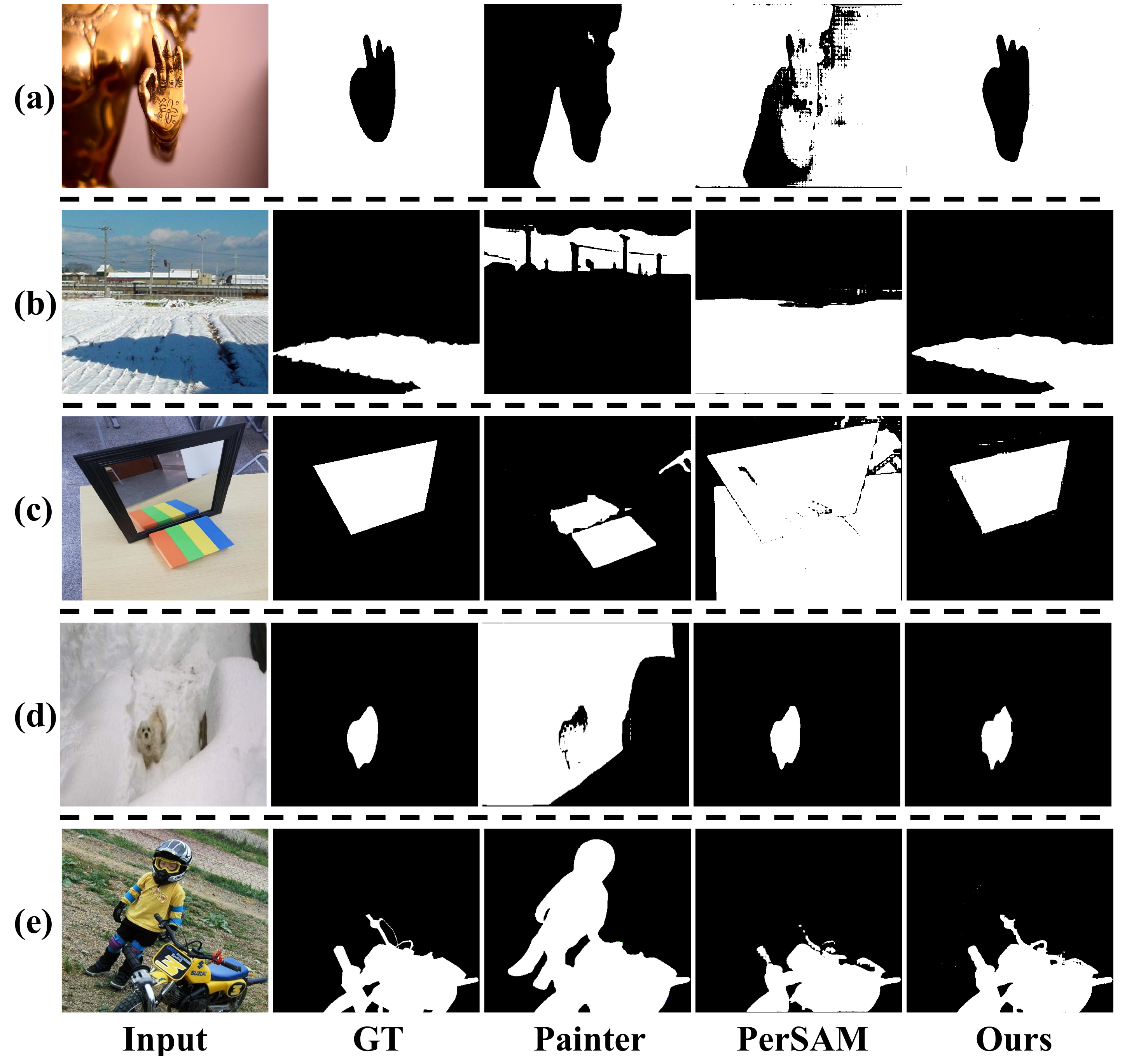}
    \caption{Qualitative comparisons between our and other methods. (a) Blur detection. (b) Shadow detection. (c) Glass detection. (d) Semantic segmentation.} 
    \label{fig:vis_seg}
    \vspace{-4mm}
\end{figure}

\subsection{Results} 
\paragraph{Quantitative results.}
Table~\ref{tab:sota_finetune} showcases the comparative performance of the proposed AlignSAM and other SOTA efficient tuning methods on four distinct segmentation tasks: blur detection, shadow detection, glass detection, and salient object detection. Notably, AlignSAM outperforms other SOTA methods in the majority of reported benchmarks across various evaluation metrics. 

To verify the effectiveness of our method in explict tasks, we further illustrate the comparison results for 20 classes on the Pascal-VOC dataset as shown in Table~\ref{tab:pascal}. Notably, AlignSAM exhibits superior performance across almost all the scenarios, achieving comparable results to the best-performing competitor for only a few categories.

\paragraph{Qualitative results.}
The comparisons of qualitative results between ours and previous SOTA algorithms on various segmentation tasks are shown in Figure~\ref{fig:vis_seg}. It can be observed that our AlignSAM is able to handle diverse challenging scenarios and produce more accurate results.

\begin{table}[t]
    \centering
    \resizebox{\linewidth}{!}{
        \begin{tabular}{c|ccccc}
        \toprule
             & Blur & Shadow & Glass & Saliency & Semantic \\
             & mIoU & mIoU & mIoU & $E_{\phi}$ & mIoU \\
        \midrule
           Implicit Branch  & \textbf{68.47} & 19.40 & \textbf{45.44} & \textbf{78.21} & 46.03 \\ 
           Explicit Branch  & 47.98 & \textbf{30.78} & 25.72 & 69.31 & \textbf{62.09} \\ 
        \bottomrule
        \end{tabular}
    }
    \caption{Ablation study results of different branches on SAM. 
    Concerning the explicit branch, the text prompts for segmentation tasks are arranged from left to right as follows: ``defocus background'', ``shadow'', ``glass'', ``salient objects'' and the category names (such as bottle, car) respectively.
    \\ 
 }
    \label{tab:srm}
    \vspace{-4mm}
\end{table}

\begin{table}[t]
\footnotesize
    \centering
        \begin{tabular}{l|cccccc}
        \toprule
            T & 1 & 5 & 10 & 15 & 20 \\
        \midrule
            Bottle & 32.59 & 40.74 & 46.42 & 48.09 & \textbf{48.70} \\ 
            Car & 42.55 & 60.15 & 65.04 & 66.13 & \textbf{66.54} \\
            Sheep & 54.69 & 67.52 & 71.16 & 73.12 & \textbf{74.00} \\ 
            Cat & 74.59 & 81.37 & 83.86 & 85.99 & \textbf{86.13} \\ 
        \bottomrule
        \end{tabular}
    \caption{Ablation study results w.r.t. iterative numbers.}
    \label{tab:point_number}
    \vspace{-4mm}
\end{table}

\subsection{Ablation Study}
\paragraph{Effect of each individual component.}
Table~\ref{tab:sota_finetune} presents a detailed abalation result of each component of AlignSAM, showcasing their individual efficacy. For example, for the blur detection task, the removal of reinforcement learning, relying solely on random action selection, resulted in an 8.72\% reduction in mIoU. Additionally, excluding the semantic recalibration module caused a 1.58\% decrease in mIoU, underscoring its role in generating precise point prompt labels. Notably, there is a significant 21.55\% decrease in mIoU when SAM failed to engage in multi-step interaction during inference, emphasizing the pivotal contribution of iterative progress. The multi-step iterative point prompt selection significantly improves mask accuracy compared to using single point prompt. Furthermore, the ablation experiments conducted on Pascal-VOC also validate the effect of AlignSAM as illustrated in Table~\ref{tab:pascal}. 

\paragraph{Effect of semantic switch.}
\vspace{-2mm}
We also examine performance differences between the implicit and explicit branches in the semantic switch in Table~\ref{tab:srm}. Abstract concept segmentation, such as saliency detection, shows superior performance when utilizing image state features only. This is due to the implicit semantic acquired from the text model is unreliable and can be detrimental to visual state features. Conversely, in concrete object segmentation, such as semantic segmentation, the utilization of leveraging hybrid vision-language features achieves better performance. The text prompt containing the category name provides valuable guidance for segmentation.

\begin{figure}[t]
    \vspace{-2mm}
    \centering    
    \includegraphics[width=0.95\linewidth]{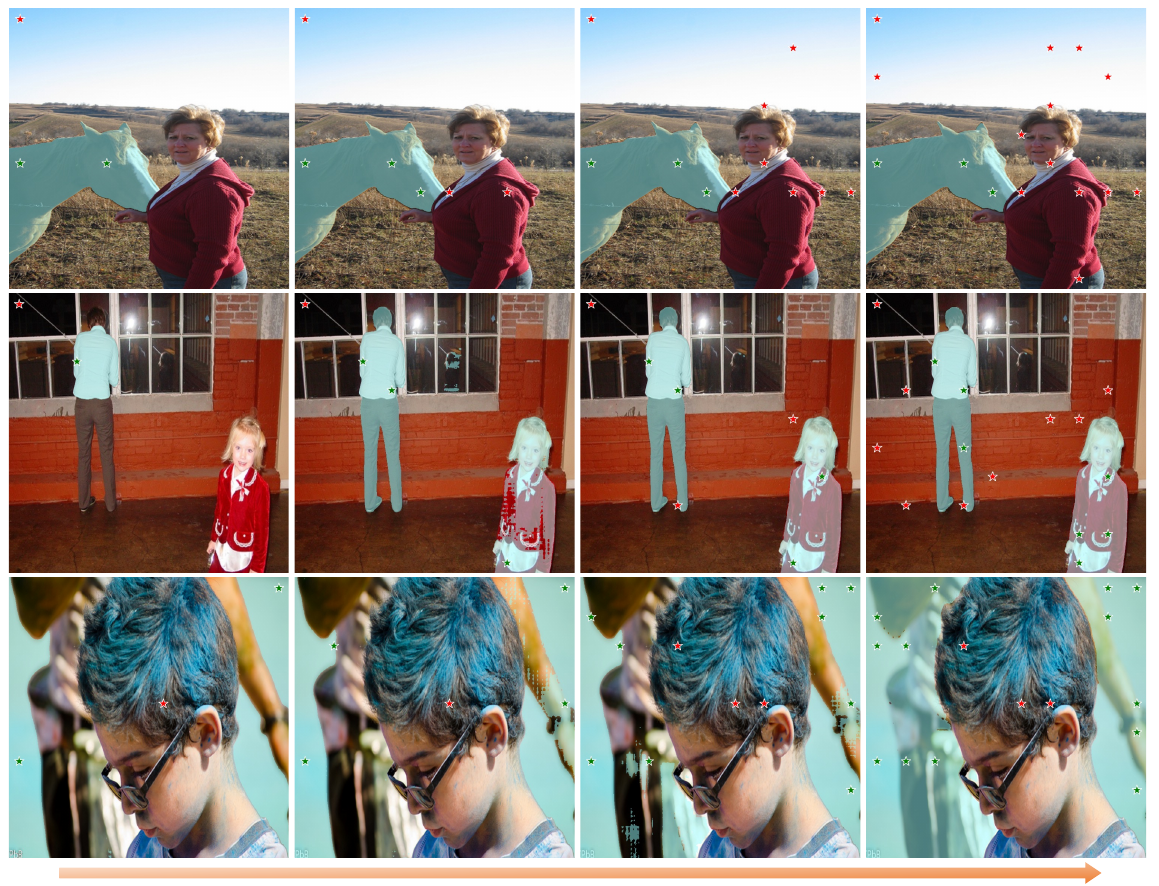}
    \caption{
    Several examples to illustrate the iterative point selection and the corresponding segmentation results. The sequence progresses from left to right, showing a gradual increase in the number of point prompts.
    } 
    \label{fig:iterative}
    \vspace{-5mm}
\end{figure}

\vspace{-2mm}

\paragraph{Effect of multi-step iterative point selection.}
We visualize the selection of iterative point prompts and report their corresponding segmentation results in Figure~\ref{fig:iterative}. 
In simple object segmentation scenarios (row 1), the model demonstrates effectiveness with only a few point prompts and more prompt points can lead to improved accuracy in the boundary of objects. However, in scenarios involving multiple objects, it requires more point prompts to segment the complete foreground areas (row 2). This limitation is also evident in low-level segmentation tasks, such as defocus blur detection (row 3). Feeding the foundation model with an adequate number of precise point prompts can significantly enhance the segmentation precision. Table~\ref{tab:point_number} further substantiates these observations, confirming the benefits of the multi-step iterative point selection process. 
\section{Conclusion}
 In this paper, we propose a novel framework for automatic prompting to align SAM to open context via reinforcement learning. 
 This unveiling of the SAM unleashes its potential across diverse downstream tasks while maintaining its adaptability intact. Initially, we devise a reinforcement learning agent to discern informative points for prompting. Subsequently, a semantic recalibration module is introduced to ensure the precise binary classification of the selected prompts. The RL agent and the recalibration module flexibly explore visual and linguistic knowledge to address the implicit and explicit semantics tasks in a unified framework. Extensive experiments conducted on diverse benchmarks confirm the efficacy of the proposed method. 

{
    \small
    \bibliographystyle{ieeenat_fullname}
    \bibliography{main}

\begin{thebibliography}{68}
\providecommand{\natexlab}[1]{#1}
\providecommand{\url}[1]{\texttt{#1}}
\expandafter\ifx\csname urlstyle\endcsname\relax
  \providecommand{\doi}[1]{doi: #1}\else
  \providecommand{\doi}{doi: \begingroup \urlstyle{rm}\Url}\fi

\bibitem[Achanta et~al.(2009)Achanta, Hemami, Estrada, and Susstrunk]{CVPR09_Fmeasure}
Radhakrishna Achanta, Sheila Hemami, Francisco Estrada, and Sabine Susstrunk.
\newblock Frequency-tuned salient region detection.
\newblock In \emph{2009 IEEE conference on computer vision and pattern recognition}, pages 1597--1604. IEEE, 2009.

\bibitem[Casanova et~al.(2020)Casanova, Pinheiro, Rostamzadeh, and Pal]{iclr20_rlal}
Arantxa Casanova, Pedro~O Pinheiro, Negar Rostamzadeh, and Christopher~J Pal.
\newblock Reinforced active learning for image segmentation.
\newblock \emph{arXiv preprint arXiv:2002.06583}, 2020.

\bibitem[Chen et~al.(2018)Chen, Zhu, Papandreou, Schroff, and Adam]{ECCV18_DeepLabV3}
Liang-Chieh Chen, Yukun Zhu, George Papandreou, Florian Schroff, and Hartwig Adam.
\newblock Encoder-decoder with atrous separable convolution for semantic image segmentation.
\newblock In \emph{Proceedings of the European conference on computer vision (ECCV)}, pages 801--818, 2018.

\bibitem[Chen et~al.(2023)Chen, Zhu, Deng, Cao, Wang, Zhang, Li, Sun, Zang, and Mao]{ICCVW23_SAM-Adapter}
Tianrun Chen, Lanyun Zhu, Chaotao Deng, Runlong Cao, Yan Wang, Shangzhan Zhang, Zejian Li, Lingyun Sun, Ying Zang, and Papa Mao.
\newblock Sam-adapter: Adapting segment anything in underperformed scenes.
\newblock In \emph{Proceedings of the IEEE/CVF International Conference on Computer Vision}, pages 3367--3375, 2023.

\bibitem[Cordts et~al.(2016)Cordts, Omran, Ramos, Rehfeld, Enzweiler, Benenson, Franke, Roth, and Schiele]{CVPR16_CITYSCAPE}
Marius Cordts, Mohamed Omran, Sebastian Ramos, Timo Rehfeld, Markus Enzweiler, Rodrigo Benenson, Uwe Franke, Stefan Roth, and Bernt Schiele.
\newblock The cityscapes dataset for semantic urban scene understanding.
\newblock In \emph{Proceedings of the IEEE conference on computer vision and pattern recognition}, pages 3213--3223, 2016.

\bibitem[Everingham et~al.(2010)Everingham, Van~Gool, Williams, Winn, and Zisserman]{IJCV10_PASCALVOC}
Mark Everingham, Luc Van~Gool, Christopher~KI Williams, John Winn, and Andrew Zisserman.
\newblock The pascal visual object classes (voc) challenge.
\newblock \emph{International journal of computer vision}, 88:\penalty0 303--338, 2010.

\bibitem[Fan et~al.(2018)Fan, Gong, Cao, Ren, Cheng, and Borji]{IJCAI18_Emeasure}
Deng-Ping Fan, Cheng Gong, Yang Cao, Bo Ren, Ming-Ming Cheng, and Ali Borji.
\newblock Enhanced-alignment measure for binary foreground map evaluation.
\newblock In \emph{Proceedings of the Twenty-Seventh International Joint Conference on Artificial Intelligence, {IJCAI-18}}, pages 698--704. International Joint Conferences on Artificial Intelligence Organization, 2018.

\bibitem[He et~al.(2021)He, Li, Cheng, Shi, Tong, Meng, Prinet, and Weng]{ICCV21_EnhanceGlassSeg}
Hao He, Xiangtai Li, Guangliang Cheng, Jianping Shi, Yunhai Tong, Gaofeng Meng, V{\'e}ronique Prinet, and LuBin Weng.
\newblock Enhanced boundary learning for glass-like object segmentation.
\newblock In \emph{Proceedings of the IEEE/CVF International Conference on Computer Vision}, pages 15859--15868, 2021.

\bibitem[He et~al.(2016)He, Zhang, Ren, and Sun]{CVPR16_ResNet}
Kaiming He, Xiangyu Zhang, Shaoqing Ren, and Jian Sun.
\newblock Deep residual learning for image recognition.
\newblock In \emph{Proceedings of the IEEE conference on computer vision and pattern recognition}, pages 770--778, 2016.

\bibitem[Houlsby et~al.(2019)Houlsby, Giurgiu, Jastrzebski, Morrone, De~Laroussilhe, Gesmundo, Attariyan, and Gelly]{ICML19_Adapter}
Neil Houlsby, Andrei Giurgiu, Stanislaw Jastrzebski, Bruna Morrone, Quentin De~Laroussilhe, Andrea Gesmundo, Mona Attariyan, and Sylvain Gelly.
\newblock Parameter-efficient transfer learning for nlp.
\newblock In \emph{International Conference on Machine Learning}, pages 2790--2799. PMLR, 2019.

\bibitem[Hu et~al.(2022)Hu, yelong shen, Wallis, Allen-Zhu, Li, Wang, Wang, and Chen]{ICLR22_LoRA}
Edward~J Hu, yelong shen, Phillip Wallis, Zeyuan Allen-Zhu, Yuanzhi Li, Shean Wang, Lu Wang, and Weizhu Chen.
\newblock Lo{RA}: Low-rank adaptation of large language models.
\newblock In \emph{International Conference on Learning Representations}, 2022.

\bibitem[Hu et~al.(2021)Hu, Wang, Fu, Jiang, Wang, and Heng]{TIP21_RevisitShadow}
Xiaowei Hu, Tianyu Wang, Chi-Wing Fu, Yitong Jiang, Qiong Wang, and Pheng-Ann Heng.
\newblock Revisiting shadow detection: A new benchmark dataset for complex world.
\newblock \emph{IEEE Transactions on Image Processing}, 30:\penalty0 1925--1934, 2021.

\bibitem[Huang et~al.(2023)Huang, Li, Chen, Huang, Chai, and Li]{huang2023divide}
Duojun Huang, Jichang Li, Weikai Chen, Junshi Huang, Zhenhua Chai, and Guanbin Li.
\newblock Divide and adapt: Active domain adaptation via customized learning.
\newblock In \emph{Proceedings of the IEEE/CVF Conference on Computer Vision and Pattern Recognition}, pages 7651--7660, 2023.

\bibitem[Jia et~al.(2021)Jia, Yang, Xia, Chen, Parekh, Pham, Le, Sung, Li, and Duerig]{icml22_align}
Chao Jia, Yinfei Yang, Ye Xia, Yi-Ting Chen, Zarana Parekh, Hieu Pham, Quoc Le, Yun-Hsuan Sung, Zhen Li, and Tom Duerig.
\newblock Scaling up visual and vision-language representation learning with noisy text supervision.
\newblock In \emph{International conference on machine learning}, pages 4904--4916. PMLR, 2021.

\bibitem[Jia et~al.(2022)Jia, Tang, Chen, Cardie, Belongie, Hariharan, and Lim]{ECCV22_VPT}
Menglin Jia, Luming Tang, Bor-Chun Chen, Claire Cardie, Serge Belongie, Bharath Hariharan, and Ser-Nam Lim.
\newblock Visual prompt tuning.
\newblock In \emph{European Conference on Computer Vision}, pages 709--727. Springer, 2022.

\bibitem[Kirillov et~al.(2023)Kirillov, Mintun, Ravi, Mao, Rolland, Gustafson, Xiao, Whitehead, Berg, Lo, Dollar, and Girshick]{ICCV23_SAM}
Alexander Kirillov, Eric Mintun, Nikhila Ravi, Hanzi Mao, Chloe Rolland, Laura Gustafson, Tete Xiao, Spencer Whitehead, Alexander~C. Berg, Wan-Yen Lo, Piotr Dollar, and Ross Girshick.
\newblock Segment anything.
\newblock In \emph{Proceedings of the IEEE/CVF International Conference on Computer Vision (ICCV)}, pages 4015--4026, 2023.

\bibitem[Li et~al.(2019{\natexlab{a}})Li, Wu, Zhang, and Huang]{a3rl}
Debang Li, Huikai Wu, Junge Zhang, and Kaiqi Huang.
\newblock Fast a3rl: Aesthetics-aware adversarial reinforcement learning for image cropping.
\newblock \emph{IEEE Transactions on Image Processing}, 28\penalty0 (10):\penalty0 5105--5120, 2019{\natexlab{a}}.

\bibitem[Li et~al.(2019{\natexlab{b}})Li, Wu, Liu, Yu, and Wong]{li2019semi}
Jichang Li, Si Wu, Cheng Liu, Zhiwen Yu, and Hau-San Wong.
\newblock Semi-supervised deep coupled ensemble learning with classification landmark exploration.
\newblock \emph{IEEE Transactions on Image Processing}, 29:\penalty0 538--550, 2019{\natexlab{b}}.

\bibitem[Li et~al.(2021)Li, Li, Shi, and Yu]{li2021cross}
Jichang Li, Guanbin Li, Yemin Shi, and Yizhou Yu.
\newblock Cross-domain adaptive clustering for semi-supervised domain adaptation.
\newblock In \emph{Proceedings of the IEEE/CVF Conference on Computer Vision and Pattern Recognition (CVPR)}, pages 2505--2514, 2021.

\bibitem[Li et~al.(2022)Li, Li, Liu, and Yu]{li2022nce}
Jichang Li, Guanbin Li, Feng Liu, and Yizhou Yu.
\newblock Neighborhood collective estimation for noisy label identification and correction.
\newblock In \emph{European Conference on Computer Vision}, pages 128--145. Springer, 2022.

\bibitem[Li et~al.(2023{\natexlab{a}})Li, Li, Cheng, Liao, and Yu]{li2024feddiv}
Jichang Li, Guanbin Li, Hui Cheng, Zicheng Liao, and Yizhou Yu.
\newblock Feddiv: Collaborative noise filtering for federated learning with noisy labels.
\newblock \emph{arXiv preprint arXiv:2312.12263}, 2023{\natexlab{a}}.

\bibitem[Li et~al.(2023{\natexlab{b}})Li, Li, and Yu]{li2023betweenness}
Jichang Li, Guanbin Li, and Yizhou Yu.
\newblock Adaptive betweenness clustering for semi-supervised domain adaptation.
\newblock \emph{IEEE Transactions on Image Processing}, 2023{\natexlab{b}}.

\bibitem[Li et~al.(2023{\natexlab{c}})Li, Li, and Yu]{li2023idm}
Jichang Li, Guanbin Li, and Yizhou Yu.
\newblock Inter-domain mixup for semi-supervised domain adaptation.
\newblock \emph{Pattern Recognition}, 2023{\natexlab{c}}.

\bibitem[Li et~al.(2024)Li, Zhang, Li, Li, Liu, Lin, and Li]{jiaming2024learning}
Jiaming Li, Jiacheng Zhang, Jichang Li, Ge Li, Si Liu, Liang Lin, and Guanbin Li.
\newblock Learning background prompts to discover implicit knowledge for open vocabulary object detection.
\newblock In \emph{Proceedings of the IEEE/CVF Conference on Computer Vision and Pattern Recognition (CVPR)}, 2024.

\bibitem[Li et~al.(2019{\natexlab{c}})Li, Wang, Li, Ma, and Wei]{li2019relation}
Luoqin Li, Jiabing Wang, Jichang Li, Qianli Ma, and Jia Wei.
\newblock Relation classification via keyword-attentive sentence mechanism and synthetic stimulation loss.
\newblock \emph{IEEE/ACM Transactions on Audio, Speech, and Language Processing}, 27\penalty0 (9):\penalty0 1392--1404, 2019{\natexlab{c}}.

\bibitem[Li et~al.(2023{\natexlab{d}})Li, Xiong, Li, and Fan]{li2023hybridvps}
Wenxue Li, Xinyu Xiong, Siying Li, and Fugui Fan.
\newblock Hybridvps: Hybrid-supervised video polyp segmentation under low-cost labels.
\newblock \emph{IEEE Signal Processing Letters}, 2023{\natexlab{d}}.

\bibitem[Li et~al.(2023{\natexlab{e}})Li, Wang, Duan, and Li]{arXiv2023_sclip}
Yi Li, Hualiang Wang, Yiqun Duan, and Xiaomeng Li.
\newblock Clip surgery for better explainability with enhancement in open-vocabulary tasks, 2023{\natexlab{e}}.

\bibitem[Li et~al.(2023{\natexlab{f}})Li, Zhang, Teng, and Lan]{arXiv23_RefSAM}
Yonglin Li, Jing Zhang, Xiao Teng, and Long Lan.
\newblock Refsam: Efficiently adapting segmenting anything model for referring video object segmentation.
\newblock \emph{arXiv preprint arXiv:2307.00997}, 2023{\natexlab{f}}.

\bibitem[Liu et~al.(2023)Liu, Zhu, Li, Chen, Wang, and Shen]{arXiv23_Matcher}
Yang Liu, Muzhi Zhu, Hengtao Li, Hao Chen, Xinlong Wang, and Chunhua Shen.
\newblock Matcher: Segment anything with one shot using all-purpose feature matching.
\newblock \emph{arXiv preprint arXiv:2305.13310}, 2023.

\bibitem[Luo et~al.(2018)Luo, Sun, Zhong, Liu, Zhang, and Wang]{iclm18_rltrack1}
Wenhan Luo, Peng Sun, Fangwei Zhong, Wei Liu, Tong Zhang, and Yizhou Wang.
\newblock End-to-end active object tracking via reinforcement learning.
\newblock In \emph{International conference on machine learning}, pages 3286--3295. PMLR, 2018.

\bibitem[Ma et~al.(2023)Ma, Wang, Liu, Lin, and Li]{ma2023enhanced}
Jie Ma, Chuan Wang, Yang Liu, Liang Lin, and Guanbin Li.
\newblock Enhanced soft label for semi-supervised semantic segmentation.
\newblock In \emph{Proceedings of the IEEE/CVF International Conference on Computer Vision}, pages 1185--1195, 2023.

\bibitem[Minaee et~al.(2021)Minaee, Boykov, Porikli, Plaza, Kehtarnavaz, and Terzopoulos]{TPAMI21_SegSurvey1}
Shervin Minaee, Yuri Boykov, Fatih Porikli, Antonio Plaza, Nasser Kehtarnavaz, and Demetri Terzopoulos.
\newblock Image segmentation using deep learning: A survey.
\newblock \emph{IEEE transactions on pattern analysis and machine intelligence}, 44\penalty0 (7):\penalty0 3523--3542, 2021.

\bibitem[Mnih et~al.(2014)Mnih, Heess, Graves, et~al.]{nips14_RAM}
Volodymyr Mnih, Nicolas Heess, Alex Graves, et~al.
\newblock Recurrent models of visual attention.
\newblock \emph{Advances in neural information processing systems}, 27, 2014.

\bibitem[Pang et~al.(2020)Pang, Zhao, Zhang, and Lu]{CVPR20_MSNet}
Youwei Pang, Xiaoqi Zhao, Lihe Zhang, and Huchuan Lu.
\newblock Multi-scale interactive network for salient object detection.
\newblock In \emph{Proceedings of the IEEE/CVF conference on computer vision and pattern recognition}, pages 9413--9422, 2020.

\bibitem[Radford et~al.(2021)Radford, Kim, Hallacy, Ramesh, Goh, Agarwal, Sastry, Askell, Mishkin, Clark, et~al.]{ICML21_CLIP}
Alec Radford, Jong~Wook Kim, Chris Hallacy, Aditya Ramesh, Gabriel Goh, Sandhini Agarwal, Girish Sastry, Amanda Askell, Pamela Mishkin, Jack Clark, et~al.
\newblock Learning transferable visual models from natural language supervision.
\newblock In \emph{International conference on machine learning}, pages 8748--8763. PMLR, 2021.

\bibitem[Reza and Kosecka(2016)]{arxiv16_rlseg}
Md~Alimoor Reza and Jana Kosecka.
\newblock Reinforcement learning for semantic segmentation in indoor scenes.
\newblock \emph{arXiv preprint arXiv:1606.01178}, 2016.

\bibitem[Rombach et~al.(2022)Rombach, Blattmann, Lorenz, Esser, and Ommer]{cvpr22_ldm}
Robin Rombach, Andreas Blattmann, Dominik Lorenz, Patrick Esser, and Bj{\"o}rn Ommer.
\newblock High-resolution image synthesis with latent diffusion models.
\newblock In \emph{Proceedings of the IEEE/CVF conference on computer vision and pattern recognition}, pages 10684--10695, 2022.

\bibitem[Schuhmann et~al.(2021)Schuhmann, Vencu, Beaumont, Kaczmarczyk, Mullis, Katta, Coombes, Jitsev, and Komatsuzaki]{laion400m}
Christoph Schuhmann, Richard Vencu, Romain Beaumont, Robert Kaczmarczyk, Clayton Mullis, Aarush Katta, Theo Coombes, Jenia Jitsev, and Aran Komatsuzaki.
\newblock Laion-400m: Open dataset of clip-filtered 400 million image-text pairs.
\newblock \emph{arXiv preprint arXiv:2111.02114}, 2021.

\bibitem[Schuhmann et~al.(2022)Schuhmann, Beaumont, Vencu, Gordon, Wightman, Cherti, Coombes, Katta, Mullis, Wortsman, et~al.]{nips22_laion5b}
Christoph Schuhmann, Romain Beaumont, Richard Vencu, Cade Gordon, Ross Wightman, Mehdi Cherti, Theo Coombes, Aarush Katta, Clayton Mullis, Mitchell Wortsman, et~al.
\newblock Laion-5b: An open large-scale dataset for training next generation image-text models.
\newblock \emph{Advances in Neural Information Processing Systems}, 35:\penalty0 25278--25294, 2022.

\bibitem[Schulman et~al.(2017)Schulman, Wolski, Dhariwal, Radford, and Klimov]{arxiv17_ppo}
John Schulman, Filip Wolski, Prafulla Dhariwal, Alec Radford, and Oleg Klimov.
\newblock Proximal policy optimization algorithms.
\newblock \emph{arXiv preprint arXiv:1707.06347}, 2017.

\bibitem[Shi et~al.(2014)Shi, Xu, and Jia]{ICCV14_CUHK}
Jianping Shi, Li Xu, and Jiaya Jia.
\newblock Discriminative blur detection features.
\newblock In \emph{Proceedings of the IEEE Conference on Computer Vision and Pattern Recognition}, pages 2965--2972, 2014.

\bibitem[Shim and Kim(2020)]{ICCAS20_GRAM}
Dongseok Shim and H~Jin Kim.
\newblock Gaussian ram: Lightweight image classification via stochastic retina-inspired glimpse and reinforcement learning.
\newblock In \emph{2020 20th International Conference on Control, Automation and Systems (ICCAS)}, pages 155--160. IEEE, 2020.

\bibitem[Simonyan and Zisserman(2015)]{ICLR15_VGG}
K Simonyan and A Zisserman.
\newblock Very deep convolutional networks for large-scale image recognition.
\newblock In \emph{International Conference on Learning Representations}, 2015.

\bibitem[Vicente et~al.(2015)Vicente, Hoai, and Samaras]{ICCV15_BER}
Tom{\'a}s F~Yago Vicente, Minh Hoai, and Dimitris Samaras.
\newblock Leave-one-out kernel optimization for shadow detection.
\newblock In \emph{Proceedings of the IEEE International Conference on Computer Vision}, pages 3388--3396, 2015.

\bibitem[Vicente et~al.(2016)Vicente, Hou, Yu, Hoai, and Samaras]{ECCV16_SBU}
Tom{\'a}s F~Yago Vicente, Le Hou, Chen-Ping Yu, Minh Hoai, and Dimitris Samaras.
\newblock Large-scale training of shadow detectors with noisily-annotated shadow examples.
\newblock In \emph{Computer Vision--ECCV 2016: 14th European Conference, Amsterdam, The Netherlands, October 11-14, 2016, Proceedings, Part VI 14}, pages 816--832. Springer, 2016.

\bibitem[Wang et~al.(2023{\natexlab{a}})Wang, Vasu, Faghri, Vemulapalli, Farajtabar, Mehta, Rastegari, Tuzel, and Pouransari]{arXiv23_SAMCLIP}
Haoxiang Wang, Pavan Kumar~Anasosalu Vasu, Fartash Faghri, Raviteja Vemulapalli, Mehrdad Farajtabar, Sachin Mehta, Mohammad Rastegari, Oncel Tuzel, and Hadi Pouransari.
\newblock Sam-clip: Merging vision foundation models towards semantic and spatial understanding.
\newblock \emph{arXiv preprint arXiv:2310.15308}, 2023{\natexlab{a}}.

\bibitem[Wang et~al.(2017)Wang, Lu, Wang, Feng, Wang, Yin, and Ruan]{CVPR2017_DUTS}
Lijun Wang, Huchuan Lu, Yifan Wang, Mengyang Feng, Dong Wang, Baocai Yin, and Xiang Ruan.
\newblock Learning to detect salient objects with image-level supervision.
\newblock In \emph{Proceedings of the IEEE Conference on Computer Vision and Pattern Recognition}, 2017.

\bibitem[Wang et~al.(2023{\natexlab{b}})Wang, Wang, Cao, Shen, and Huang]{CVPR23_Painter}
Xinlong Wang, Wen Wang, Yue Cao, Chunhua Shen, and Tiejun Huang.
\newblock Images speak in images: A generalist painter for in-context visual learning.
\newblock In \emph{Proceedings of the IEEE/CVF Conference on Computer Vision and Pattern Recognition}, pages 6830--6839, 2023{\natexlab{b}}.

\bibitem[Wang et~al.(2023{\natexlab{c}})Wang, Zhang, Cao, Wang, Shen, and Huang]{ICCV23_SegGPT}
Xinlong Wang, Xiaosong Zhang, Yue Cao, Wen Wang, Chunhua Shen, and Tiejun Huang.
\newblock Seggpt: Towards segmenting everything in context.
\newblock In \emph{Proceedings of the IEEE/CVF International Conference on Computer Vision}, pages 1130--1140, 2023{\natexlab{c}}.

\bibitem[Wang et~al.(2023{\natexlab{d}})Wang, Wang, Fan, Wang, and He]{CVPR23_MESOD}
Yi Wang, Ruili Wang, Xin Fan, Tianzhu Wang, and Xiangjian He.
\newblock Pixels, regions, and objects: Multiple enhancement for salient object detection.
\newblock In \emph{Proceedings of the IEEE/CVF Conference on Computer Vision and Pattern Recognition}, pages 10031--10040, 2023{\natexlab{d}}.

\bibitem[Wu et~al.(2023)Wu, Fu, Fang, Liu, Wang, Xu, Jin, and Arbel]{arXiv23_MSA}
Junde Wu, Rao Fu, Huihui Fang, Yuanpei Liu, Zhaowei Wang, Yanwu Xu, Yueming Jin, and Tal Arbel.
\newblock Medical sam adapter: Adapting segment anything model for medical image segmentation.
\newblock \emph{arXiv preprint arXiv:2304.12620}, 2023.

\bibitem[Wu et~al.(2019{\natexlab{a}})Wu, Deng, Li, Li, Yu, and Wong]{wu2019enhancing}
Si Wu, Guangchang Deng, Jichang Li, Rui Li, Zhiwen Yu, and Hau-San Wong.
\newblock Enhancing triplegan for semi-supervised conditional instance synthesis and classification.
\newblock In \emph{Proceedings of the IEEE/CVF Conference on Computer Vision and Pattern Recognition}, pages 10091--10100, 2019{\natexlab{a}}.

\bibitem[Wu et~al.(2019{\natexlab{b}})Wu, Li, Liu, Yu, and Wong]{wu2019mutual}
Si Wu, Jichang Li, Cheng Liu, Zhiwen Yu, and Hau-San Wong.
\newblock Mutual learning of complementary networks via residual correction for improving semi-supervised classification.
\newblock In \emph{Proceedings of the IEEE/CVF conference on computer vision and pattern recognition}, pages 6500--6509, 2019{\natexlab{b}}.

\bibitem[Xiong et~al.(2023{\natexlab{a}})Xiong, Li, and Li]{xiong2023unpaired}
Xinyu Xiong, Siying Li, and Guanbin Li.
\newblock Unpaired image-to-image translation based domain adaptation for polyp segmentation.
\newblock In \emph{International Symposium on Biomedical Imaging}, pages 1--5. IEEE, 2023{\natexlab{a}}.

\bibitem[Xiong et~al.(2023{\natexlab{b}})Xiong, Wang, Li, and Li]{MLMI23_MammoSAM}
Xinyu Xiong, Churan Wang, Wenxue Li, and Guanbin Li.
\newblock Mammo-sam: Adapting foundation segment anything model for automatic breast mass segmentation in whole mammograms.
\newblock In \emph{International Workshop on Machine Learning in Medical Imaging}, pages 176--185. Springer, 2023{\natexlab{b}}.

\bibitem[Yang et~al.(2019)Yang, Mei, Xu, Wei, Yin, and Lau]{ICCV19_MSD}
Xin Yang, Haiyang Mei, Ke Xu, Xiaopeng Wei, Baocai Yin, and Rynson~WH Lau.
\newblock Where is my mirror?
\newblock In \emph{Proceedings of the IEEE/CVF International Conference on Computer Vision}, pages 8809--8818, 2019.

\bibitem[Yu et~al.(2022)Yu, Mei, Dong, Wei, Zhu, Wang, and Yang]{TIP22_ProgreGlassSeg}
Letian Yu, Haiyang Mei, Wen Dong, Ziqi Wei, Li Zhu, Yuxin Wang, and Xin Yang.
\newblock Progressive glass segmentation.
\newblock \emph{IEEE Transactions on Image Processing}, 31:\penalty0 2920--2933, 2022.

\bibitem[Zhang et~al.(2017)Zhang, Maei, Wang, and Wang]{arxiv17_rltrack2}
Da Zhang, Hamid Maei, Xin Wang, and Yuan-Fang Wang.
\newblock Deep reinforcement learning for visual object tracking in videos.
\newblock \emph{arXiv preprint arXiv:1701.08936}, 2017.

\bibitem[Zhang and Liu(2023)]{arXiv23_CustomSAM}
Kaidong Zhang and Dong Liu.
\newblock Customized segment anything model for medical image segmentation.
\newblock \emph{arXiv preprint arXiv:2304.13785}, 2023.

\bibitem[Zhang et~al.(2023)Zhang, Jiang, Guo, Yan, Pan, Dong, Gao, and Li]{arXiv23_PerSAM}
Renrui Zhang, Zhengkai Jiang, Ziyu Guo, Shilin Yan, Junting Pan, Hao Dong, Peng Gao, and Hongsheng Li.
\newblock Personalize segment anything model with one shot.
\newblock \emph{arXiv preprint arXiv:2305.03048}, 2023.

\bibitem[Zhang et~al.(2024{\natexlab{a}})Zhang, Chen, Wang, and Yang]{10097456}
Zhicheng Zhang, Song Chen, Zichuan Wang, and Jufeng Yang.
\newblock Planeseg: Building a plug-in for boosting planar region segmentation.
\newblock \emph{IEEE Transactions on Neural Networks and Learning Systems}, 1\penalty0 (1):\penalty0 1--15, 2024{\natexlab{a}}.

\bibitem[Zhang et~al.(2024{\natexlab{b}})Zhang, Hu, Cheng, Paudel, and Yang]{zhang2024distribution}
Zhicheng Zhang, Junyao Hu, Wentao Cheng, Danda Paudel, and Jufeng Yang.
\newblock Extdm: Distribution extrapolation diffusion model for video prediction.
\newblock In \emph{Proceedings of the IEEE/CVF Conference on Computer Vision and Pattern Recognition (CVPR)}, 2024{\natexlab{b}}.

\bibitem[Zhang et~al.(2024{\natexlab{c}})Zhang, Zhao, Park, and Yang]{zhang2024masked}
Zhicheng Zhang, Pancheng Zhao, Eunil Park, and Jufeng Yang.
\newblock Mart: Masked affective representation learning via masked temporal distribution distillation.
\newblock In \emph{Proceedings of the IEEE/CVF Conference on Computer Vision and Pattern Recognition (CVPR)}, 2024{\natexlab{c}}.

\bibitem[Zhao et~al.(2019)Zhao, Liu, Fan, Cao, Yang, and Cheng]{ICCV19_EGNet}
Jia-Xing Zhao, Jiang-Jiang Liu, Deng-Ping Fan, Yang Cao, Jufeng Yang, and Ming-Ming Cheng.
\newblock Egnet: Edge guidance network for salient object detection.
\newblock In \emph{Proceedings of the IEEE/CVF international conference on computer vision}, pages 8779--8788, 2019.

\bibitem[Zhao et~al.(2020)Zhao, Qian, Zhang, Li, Wei, Liu, and Pan]{ICDM20_Dice}
Rongjian Zhao, Buyue Qian, Xianli Zhang, Yang Li, Rong Wei, Yang Liu, and Yinggang Pan.
\newblock Rethinking dice loss for medical image segmentation.
\newblock In \emph{2020 IEEE International Conference on Data Mining (ICDM)}, pages 851--860. IEEE, 2020.

\bibitem[Zheng et~al.(2019)Zheng, Qiao, Cao, and Lau]{CVPR19_DISTRACTION}
Quanlong Zheng, Xiaotian Qiao, Ying Cao, and Rynson~WH Lau.
\newblock Distraction-aware shadow detection.
\newblock In \emph{Proceedings of the IEEE/CVF Conference on Computer Vision and Pattern Recognition}, pages 5167--5176, 2019.

\bibitem[Zhu et~al.(2021)Zhu, Xu, Ke, and Lau]{iccv21_fdrnet}
Lei Zhu, Ke Xu, Zhanghan Ke, and Rynson~WH Lau.
\newblock Mitigating intensity bias in shadow detection via feature decomposition and reweighting.
\newblock In \emph{Proceedings of the IEEE/CVF International Conference on Computer Vision}, pages 4702--4711, 2021.

\bibitem[Zou et~al.(2023)Zou, Yang, Zhang, Li, Li, Gao, and Lee]{arXiv23_SEEM}
Xueyan Zou, Jianwei Yang, Hao Zhang, Feng Li, Linjie Li, Jianfeng Gao, and Yong~Jae Lee.
\newblock Segment everything everywhere all at once.
\newblock \emph{arXiv preprint arXiv:2304.06718}, 2023.

\end{thebibliography}
}

\newpage
\appendix
\section*{Supplementary Materials}
\noindent
We present additional implementation details and analysis of our proposed method AlignSAM in this supplementary material.

\section{Implementation Details}
We illustrate the textual prompts for the implentation of explict branch on different datasets in Table~\ref{tab:prompt}. The ``Text Prompt'' in the last column is utilized as textual input of the CLIP-Surgery Model. 

\begin{table}[h]
\centering
\resizebox{0.48\textwidth}{!}
{
\begin{tabular}{l|cc} 
\hline
Datasets & Target Foreground & Text Prompt \\ 
\hline
CUHK~\cite{ICCV14_CUHK} & Defocus background  &    defocus background     \\
\hline
SBU~\cite{ECCV16_SBU} & Shadow of any object  &  shadow   \\
\hline
MSD~\cite{ICCV19_MSD} &  Mirror face &  glass    \\
\hline
DUTS~\cite{CVPR2017_DUTS} & Visually salient objects & saliency object \\
\hline
Pascal-VOC~\cite{IJCV10_PASCALVOC} & Common categories & aeroplane/ bottle/...
\\
\hline 
\end{tabular}
}
\caption{
Summary of textual prompts for different datasets. 
} 
\label{tab:prompt}
\end{table}
\vspace{-10pt}

\section{Comparisons of Tuning Paradigms}
This paper introduces a unified framework based on reinforcement learning to enable effective and efficient prompting for the vision foundation model SAM, without the need to access the parameters of the backbone. We here report the comparison results of our and existing paradigms for adapting SAM into downstream tasks, in terms of five desirable properties, including frozen backbone ($\mathbf{F}$), automatic inference ($\mathbf{A}$), gradient-free ($\mathbf{G}$), source-free ($\mathbf{S}$), and interpretability ($\mathbf{I}$). 
Specifically, freezing the parameters ($\mathbf{F}$) of the foundation model is required to alleviate the burden of training costs due to the giant scale of the backbone model. ``$\mathbf{A}$'' means automatic inference without additional guidance in the testing phase. Gradient-free methods ($\mathbf{G}$) do not require gradient information from the intermediate layer of the foundation model, which can be expensive to compute during training. Source-free ($\mathbf{S}$) methods do not require training (reference) samples in the testing phase which may be inaccessible due to privacy issues. Interpretability ($\mathbf{I}$) implies that the prediction results are explicitly and strongly connected to the provided prompts for the foundation model. 
As depicted in Table~\ref{tab:tuning_way}, our proposed \shortname{} combines various desirable properties that can be utilized for diverse downstream tasks. 

\begin{table}[ht]
\centering
\begin{tabular}{l|ccccc} 
\hline
Methods & $\mathbf{F}$ & $\mathbf{A}$ & $\mathbf{G}$ & $\mathbf{S}$ & $\mathbf{I}$ \\ 
\hline
Manual Prompting & \cmark  & \xmark & \cmark &  \cmark & \cmark       \\
Full Fine-tuning & \xmark  & \cmark & \xmark & \cmark & \xmark  \\
Adapters / LoRAs~\cite{ICCVW23_SAM-Adapter,arXiv23_CustomSAM} & \cmark & \cmark & \cmark & \cmark & \xmark                     \\
In-Context Learning~\cite{arXiv23_PerSAM} & \cmark  & \cmark & \cmark &    \xmark & \cmark  \\
\hline
Ours & \cmark &  \cmark & \cmark & \cmark & \cmark \\
\hline 
\end{tabular}
\caption{
Comparison results of our and existing paradigms for adapting SAM into downstream tasks, in terms of five desirable properties.
} 
\label{tab:tuning_way}
\end{table}

\section{Analysis of Prompt Label} 
At each timestep, the RL agent chooses a prompt position from the action space and queries its corresponding label (foreground or background) from the output of the SRM module. Theoretically, all the selected points can be considered positive foreground points since the agent is trained to prioritize foreground areas. However, the RL agent can only identify the target region at a coarse level due to the limited space of actions, leading to potentially unreliable labeling. Alternatively, the prompt label can be queried from the last prediction of SAM or the Vision-Language similarity map. To verify the efficacy of the reference mask for querying the prompt's label, we replace the SRM module by the above-mentioned variants. As shown in Table~\ref{tab:ab_plab}, our method utilizing the prediction of the SRM module as the label source consistently outperforms the other alternatives, demonstrating the superiority of the prompt labeling strategy. 

\begin{table}[h]
\footnotesize
\centering
	\begin{tabular}{l|c|cccc} 
	\toprule
  \makecell*[l]{Method} &\makecell*[c]{mIoU}
        &\makecell*[c]{Bottle}  &\makecell*[c]{Car}  &\makecell*[c]{Sheep} &\makecell*[c]{Cat} \\

		\cmidrule(lr){1-1} \cmidrule(lr){2-2}  
  \cmidrule(lr){3-6}
        RL+SCLIP & 58.24 & 45.64 & 62.39 & 64.40 & 85.20   \\  
        RL+LAST & 24.20 & 21.59 & 24.40 & 33.09 & 40.24 \\
        RL only & 27.73 & 19.78 & 27.42 & 30.33 & 39.29 \\
        \cmidrule(lr){1-6}
        \bf RL+SRM(Ours) & \textbf{62.09} & \textbf{48.09} & \textbf{66.13} & \textbf{73.12} & \textbf{85.99} \\
	\bottomrule
	\end{tabular}
\caption{
Ablation study results of the proposed prompt labeling module SRM. The performance results are calculated by averaging IoU ($\%$) of all the
categories in Pascal-VOC 2012. ``RL+X'' denotes utilizing the trained RL agent to perform prompt selection and query the label of prompt from ``X''. 
}
 \label{tab:ab_plab}
\end{table}

\begin{table*}[ht]
\centering
\footnotesize
{
\begin{tabular}{l|cc|cc|cc|cc|cc}
\toprule
\multirow{3}{*}{Method} & \multicolumn{2}{c|}{\textbf{Blur}} & 
\multicolumn{2}{c|}{\textbf{Shadow}}  & \multicolumn{2}{c|}{\textbf{Glass}} & \multicolumn{2}{c|}{\textbf{Saliency}} & \multicolumn{2}{c}{\textbf{Semantic}} \\
      &  \multicolumn{2}{c|}{CUHK~\cite{ICCV14_CUHK}} & \multicolumn{2}{c|}{SBU~\cite{ECCV16_SBU}}  & \multicolumn{2}{c|}{MSD~\cite{ICCV19_MSD}} & \multicolumn{2}{c|}{DUTS~\cite{CVPR2017_DUTS}} & \multicolumn{2}{c}{Pascal VOC~\cite{IJCV10_PASCALVOC}}  \\ 
     &  mIoU $\uparrow$ &  $\textit{FR} $   & mIoU $\uparrow$ & \textit{FR}  & mIoU $\uparrow$ &  \textit{FR} &  $E_\phi \uparrow$ & \textit{FR} &  mIoU $\uparrow$ &  $\textit{FR} $ 
          \\ \hline  
Ours-w/o RL & 59.75 &  63.80 & 25.62 & 18.30 & 33.41 &  22.80 & 74.19 & 14.60 & 54.06 & 23.20 \\ 
\textbf{Ours} & \textbf{68.47} & 70.90  & \textbf{30.78} & 24.80 & \textbf{45.44} & 43.70 & \textbf{78.21} & 32.90 & \textbf{62.09} & 36.70 \\

\bottomrule
\end{tabular}}
\caption{
Ablation study results of the RL policy on different segmentation tasks. ``Ours-w/o RL'' denotes the degraded variant of the proposed approach where the RL policy is replaced with random selection for the action decision. The best performance among all approaches is highlighted in \textbf{blod}. 
}
\label{tab:RL_fgr}
\vspace{-10pt}
\end{table*}

\begin{figure}[t]
    \vspace{-2mm}
    \centering    
    \includegraphics[width=0.95\linewidth]{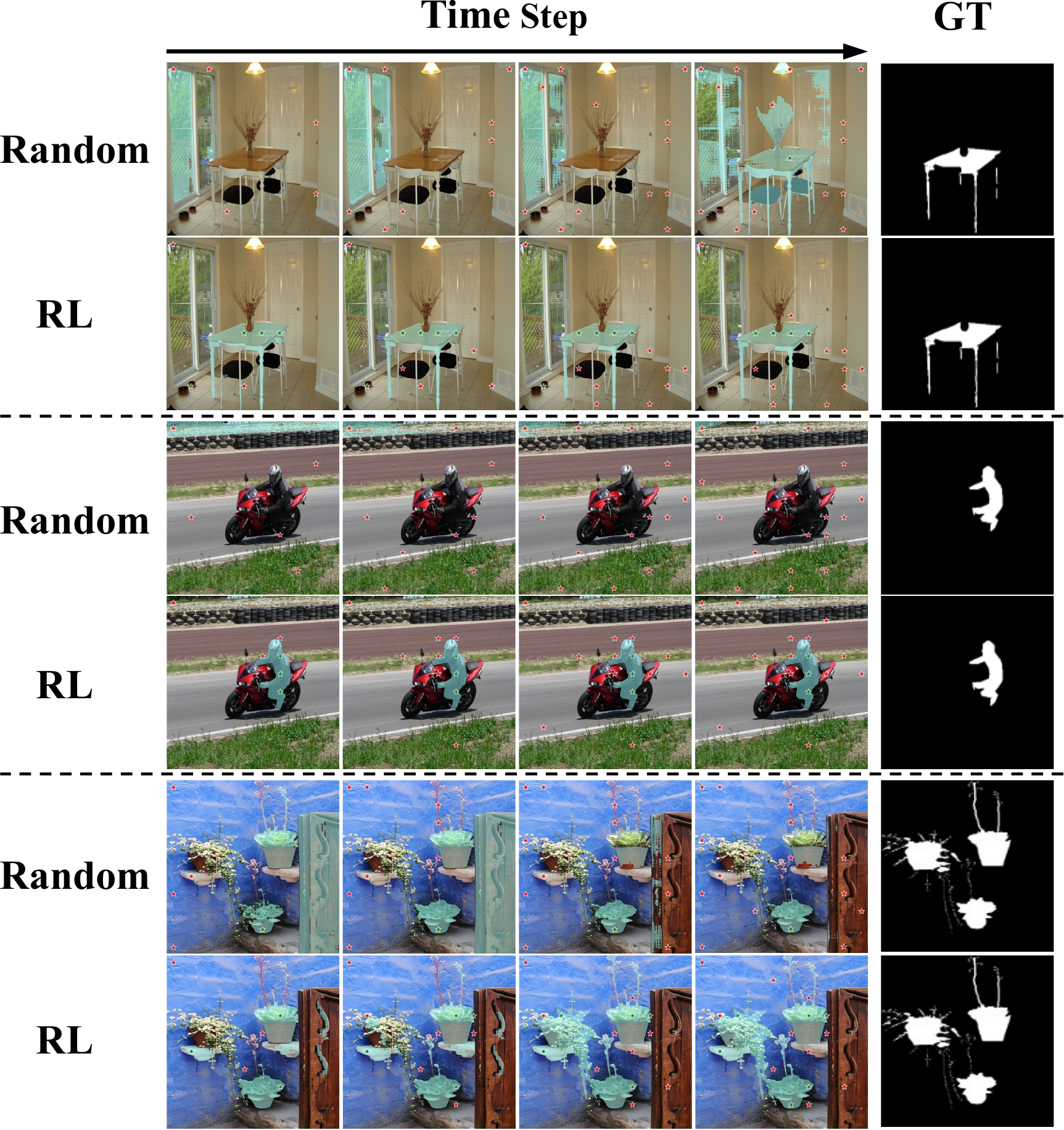}
    \caption{
    Representative examples to illustrate the iterative point selection and the corresponding segmentation results. The sequence progresses from left to right, showing a gradual increase in the number of point prompts. The 
    green and red stars denote the positive and negative point prompts respectively. The input image is covered by the segmentation mask shown in a light green color. ``GT'' denotes the ground truth mask of the input image. 
    } 
    \label{fig:iterative_fgr}
    \vspace{-20pt}
\label{fig:RL_fgr}
\end{figure}

\section{Analysis of RL Policy}
To investigate the strategy learned by the reinforcement learning agent after training, we assess the disparity in action selection between reinforcement learning strategy and random sampling. We propose a Foreground Rate (FR) to measure the ability to track the target of interest, which can be formulated as follows: 
\begin{equation}\label{class_centroid}
\textit{FR} = \sum_{t\in [1, T]} \mathds{1}{\{ G_I(a_t)=1\}} / T, 
\end{equation} 
where $G_I(a_t)$ means querying the label of the chosen position from the ground truth mask for sample $I$. For both the RL and random strategies, we set the number of interaction rounds $T$ to 15 and report $\textit{FR}$ averaged on all the testing samples in each dataset. As shown in Table~\ref{tab:RL_fgr}, the $\textit{FR}$ of the RL strategy significantly surpasses that of random selection in all the reported scenarios, indicating the high-value estimation of actions in the target area by the RL network. As illustrated in Figure~\ref{fig:RL_fgr}, our RL agent exhibits a greater inclination for selecting point prompts in the vicinity of the target area compared to random selection. The iterative segmentation process highlights the advantages of employing the reinforcement learning strategy in multiple aspects. First, with a limited prompting budget, the RL agent can proficiently capture the area of the target of interest, thereby facilitating the segmentation of SAM. Secondly, in scenarios with multiple disjointed target regions within an image, utilizing the RL agent for prompt selection effectively prevents the omission of target regions. In summary, our RL agent consistently selects more points surrounding the target area than random selection in diversified scenarios, thereby unlocking the potential for progressive segmentation refinement. 

\vspace{-10pt}
\section{Analysis of Training Samples} 
To validate the robustness of \shortname{}, we compare it with other competitive state-of-the-art (SOTA) methods across various budgets of training samples. For each dataset, the training samples are randomly sampled from the training set and shared among different methods to ensure a fair comparison. As depicted in Table~\ref{tab:diff_shot}, \shortname{} consistently outperforms the second-best SOTA method in most scenarios. This implies that the RL agent and the SRM module can synergistically align SAM to the segmentation objective even with a limited number of reference samples. 

\begin{table}[t]
\centering
\footnotesize
\resizebox{0.48\textwidth}{!}
{
\begin{tabular}{l|cc|cc|cc}
\toprule
\multirow{3}{*}{Method} & \multicolumn{2}{c|}{\textbf{Blur}} & 
\multicolumn{2}{c|}{\textbf{Glass}} & \multicolumn{2}{c}{\textbf{Saliency}} \\
      &  \multicolumn{2}{c|}{CUHK~\cite{ICCV14_CUHK}} & \multicolumn{2}{c|}{MSD~\cite{ICCV19_MSD}}  & \multicolumn{2}{c}{DUTS~\cite{CVPR2017_DUTS}}\\ 
      & 5-shot &  20-shot   & 5-shot &  20-shot  & 5-shot &  20-shot 
          \\ \hline 
PerSAM~\cite{arXiv23_PerSAM} & 53.21 & 54.69 & 31.03 & 29.50 & \textbf{35.92} & 40.50 
\\ 
\textbf{Ours} & \textbf{68.89} & \textbf{62.99} & \textbf{31.19} & \textbf{38.78} & 33.05 & \textbf{44.82}  \\ 
\bottomrule
\end{tabular}}
\vspace{-10pt}
\caption{
Comparison results of our AlignSAM and PerSAM under different numbers of training samples. The reported results denote the performance of the models evaluated by mean IoU. The best performance among all approaches is highlighted in \textbf{blod}. The second-best competitor PerSAM~\cite{arXiv23_PerSAM} is selected as the representative of SOTA methods. }
\label{tab:diff_shot}
\vspace{-2mm}
\end{table}

\begin{figure}[t]
\centering
    \hspace{-3mm}
    \subfloat[\ \ \ Sensitivity to $T$]{
    \label{fig:sense-m}
    \includegraphics[width=0.233\textwidth]{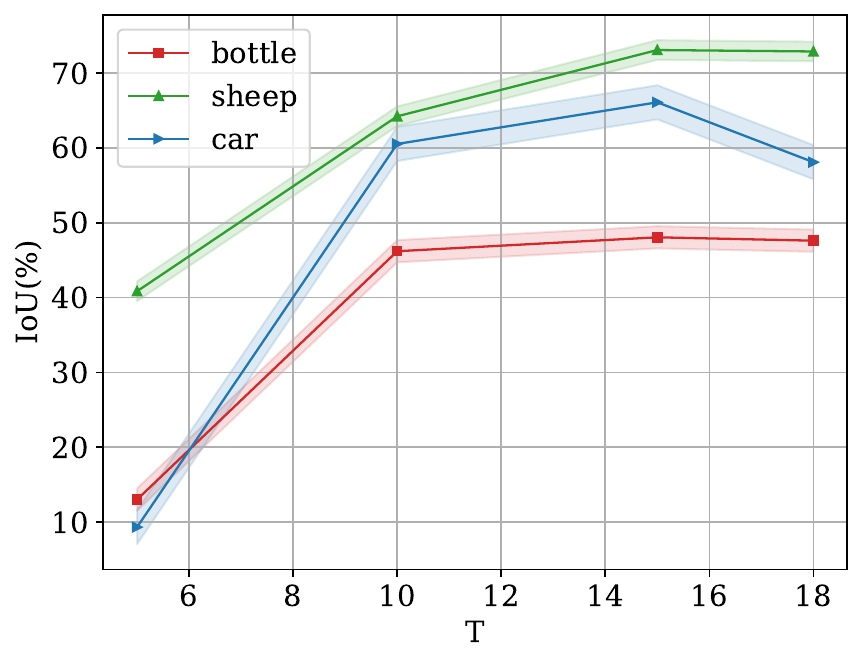}
    }
    \hspace{-3mm}
    \subfloat[\ \ \ Sensitivity to $E$]{
    \label{fig:sense-lambda}
    \includegraphics[width=0.233\textwidth]{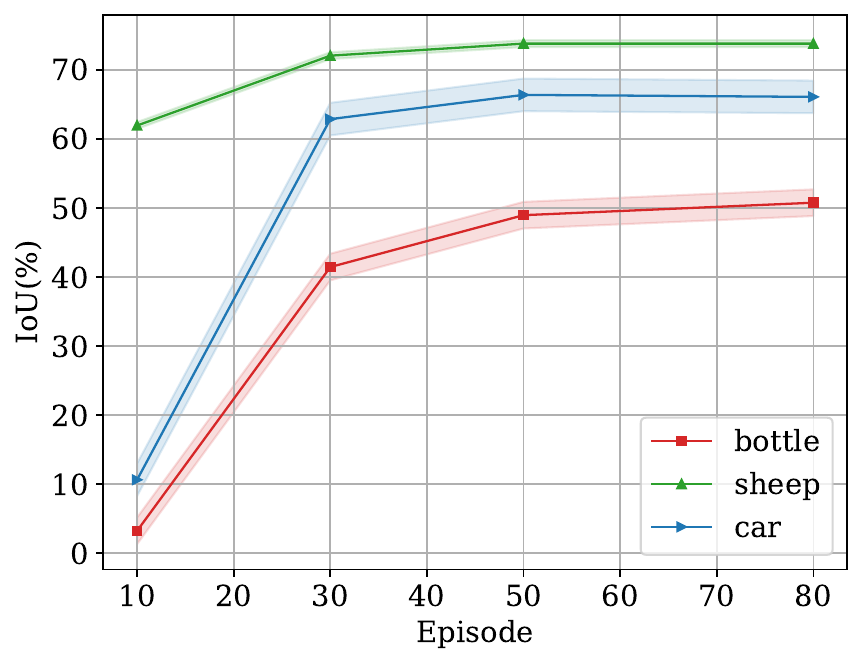}
    }
    \vspace{-10pt}
    \caption{Hyper-parameter sensitivity to $T$ and $E$ of \shortname{}.}
    \label{fig:sensi}
    \vspace{-15pt}
\end{figure}

\vspace{-5pt}
\section{Hyper-Parameter Sensitivity}
We further carry out investigations to check the sensitivity of the proposed approach to the key hyper-parameters $E$ and $T$. These experiments are conducted under three scenarios with varying degrees of segmentation difficulty. In Figure~\ref{fig:sensi}, we show the model performance of \shortname{} when $T$ and $E$ are respectively set to $\{5, 10, 15, 18\}$ and $\{10, 30, 50, 80\}$. 
As illustrated, the model trained with $T=15$ and $E=50$ exhibits notably superior and stable performance compared to other configurations. This emphasizes the efficacy of setting both to 15 and 50, respectively, as a good choice for the implementation. 
\vspace{-10pt}

\end{document}